\documentclass[final,5p,times,twocolumn]{elsarticle}

\usepackage[font=small,labelfont=bf]{caption}
\usepackage{amsmath,amssymb}
\usepackage{xcolor}
\usepackage{url}
\usepackage[colorlinks=true,allcolors=blue]{hyperref}
\usepackage{booktabs}
\usepackage{multirow}
\usepackage{array}
\usepackage{adjustbox}
\usepackage{makecell}
\usepackage{subcaption}
\usepackage{dblfloatfix}
\biboptions{authoryear}
\newcommand{\std}[1]{\,{\scriptsize$\pm$#1}}
\begin{document}

    \begin{frontmatter}

        \title{GALA-3D: Self-Supervised Global and Local Anatomical Pretraining for Label-Efficient 3D Medical Image Segmentation}

        \author[1,2]{Aloys Portafaix\corref{cor1}}
        \cortext[cor1]{Corresponding author: \href{mailto:aloys.portafaix@polymtl.ca}{aloys.portafaix@polymtl.ca}}
        \author[2,3]{Laurent L\'etourneau-Guillon}
        \author[1,2]{Samuel Kadoury}

        \address[1]{Polytechnique Montr\'eal, 2500 chemin de Polytechnique, Montr\'eal, QC H3T 1J4, Canada}
        \address[2]{Centre de recherche du Centre hospitalier de l'Universit\'e de Montr\'eal (CRCHUM), 900 rue Saint-Denis, Montr\'eal, QC H2X 0A9, Canada}
        \address[3]{Department of Radiology, Radiation Oncology, and Nuclear Medicine, Universit\'e de Montr\'eal, 2900 boulevard \'Edouard-Montpetit, Montr\'eal, QC, H3T 1A4, Canada}

        \begin{abstract}
            Annotating three-dimensional medical images for segmentation is labour-intensive, and self-supervised pretraining on unlabelled images can reduce this burden. However, existing objectives either reconstruct voxel intensities, favouring local appearance over anatomical semantics, or rely on momentum teachers and other anti-collapse mechanisms that add training complexity. We propose GALA-3D, a teacher-free framework that extends the joint-embedding objective of LeJEPA to 3D images with two components. First, anatomy-aware multi-view sampling draws global and local views from a shared foreground-anchored region, so that positive views depict related anatomy. Second, a dense branch predicts the latent representations of masked patches from their visible context. Because teacher-free dense prediction is prone to collapse, we stabilize it with a patch-level isotropic Gaussian regularizer. The dense branch adds only 5\% to the computational cost. Pretrained on a single GPU with 25,414 abdominal CT images disjoint from all evaluation datasets, GALA-3D obtains the highest mean Dice among state-of-the-art methods on seven public CT segmentation datasets at every label budget, and GALA-B exceeds the strongest external model by 2.2 to 3.1 Dice points from one labelled case to 10\% of the labels. A masked autoencoder pretrained on the same images with the same encoders trails GALA-3D by 2.3 to 5.8 points at these budgets, which attributes the gain over masked reconstruction to the pretraining framework rather than to the data. Pretrained on a head CT cohort, GALA-3D also becomes the best method on an intracerebral haemorrhage dataset with limited labels. Code and weights will be released at \url{https://github.com/aloyspx/GALA-3D}.
        \end{abstract}


    \end{frontmatter}

    \section{Introduction}
    \label{sec:introduction}

    Despite the important progress of deep learning in medical image analysis, its success continues to rely on large-scale expert annotations, which are costly and time-consuming to obtain \citep{surverydlmed}. This dependency is particularly limiting for three-dimensional (3D) images, where delineating anatomical structures throughout an entire volume is laborious, expensive, and difficult to scale \citep{burden_annotation}. Self-supervised learning (SSL) offers a promising alternative: a model can first be trained on large collections of unlabelled images, which are ubiquitous in hospital archives and increasingly available in the public space, and subsequently adapted to segmentation and other downstream tasks using substantially fewer annotations \citep{learningwithless}. However, for this promise to be realized, the pretraining objective must yield representations that are both semantically useful and practical to train at scale.

    Masked image modelling (MIM) provides an appealing foundation for volumetric pretraining because of its simplicity and efficiency, as expensive encoder operations can be restricted to a small subset of visible input tokens \citep{he_masked_2021,hi_end_mae}. However, conventional masked autoencoders (MAEs) are optimized to reconstruct low-level image content, which favours local appearance over anatomically meaningful features. Contrastive approaches instead teach higher-level invariances by enforcing agreement between augmented views \citep{chen_simple_2020,vox2vec}, but their negative pairs can be undesirable in medical imaging, where images or regions containing similar anatomy may be pushed apart \citep{boosting_contrastive_learning,anatomy_aware_contrastive}. Non-contrastive methods remove explicit negatives, but they typically rely on a momentum teacher or other asymmetric mechanisms to avoid representational collapse \citep{dinov3,3dino}, which introduce additional hyperparameters, such as the teacher momentum and temperature. This motivates a simpler pretraining framework that jointly learns globally consistent and locally informative anatomical representations.

    LeJEPA \citep{lejepa} provides an attractive starting point towards this objective. It learns consistent representations across related views and regularizes their distribution with Sketched Isotropic Gaussian Regularization (SIGReg), which yields a compact joint-embedding objective without negative pairs or a momentum teacher. However, applying this objective directly to volumetric medical images raises two challenges. First, a 3D image may span a large anatomical field of view, such that independently sampled crops can depict entirely different anatomical regions and therefore form inappropriate positive views. Second, an image-level objective alone does not explicitly encourage the spatially resolved anatomical information required by dense downstream tasks such as segmentation.

    We propose GALA-3D, \textbf{G}lobal \textbf{A}nd \textbf{L}ocal \textbf{A}natomical representation learning for self-supervised \textbf{3D} medical image pretraining, named after its global and local anatomical views. GALA-3D extends LeJEPA with one component for each challenge. First, global and local views are drawn from a common foreground-anchored parent region rather than independently from the whole image, so that positive views remain anatomically related while covering complementary fields of view. Second, a dense branch infers the latent representations of masked patches from their visible context, with a visible-token encoder and a cross-attention predictor. Because its targets are latent representations produced by a clean pass of the same encoder, rather than voxel intensities or the outputs of a teacher network, this branch encourages spatially resolved anatomical representations that complement the global objective. However, we find that teacher-free dense latent prediction collapses without a dedicated regularizer. We therefore introduce a patch-level SIGReg, which evaluates slots of sampled patch embeddings across the independent images of the batch and thereby enables stable training of the dense branch. Encoding only the few visible tokens of each masked view, the dense branch adds 5\% to the operations of a training step. We evaluate GALA-3D on in-domain abdominal computed tomography (CT), out-of-domain head CT, and brain magnetic resonance imaging (MRI) segmentation across four label budgets, and compare it with eight publicly released pretrained medical models. We further pretrain GALA-3D on a head CT cohort to test whether its out-of-domain results are limited by the method or by the pretraining data.

    The contributions of this work are as follows:

    \begin{itemize}
        \item \textbf{Anatomy-aware Multi-View Pipeline:} We introduce a view-generation strategy that constructs global and local positive views from a shared foreground-anchored parent region, so that every positive view depicts related anatomy while covering a different field of view.
        \item \textbf{Global and Dense Latent Objective:} We extend LeJEPA to 3D medical images with teacher-free masked latent-patch prediction. We find that such a prediction collapses without a dedicated regularizer, and we introduce a patch-level SIGReg that evaluates slots of sampled patch embeddings across the independent images of the batch. We show that this regularizer is necessary for stable dense training and that the global and dense branches are complementary.
        \item \textbf{Label-efficient Transfer:} On seven CT segmentation datasets evaluated with five-fold cross-validation at four label budgets, GALA-3D obtains the highest mean Dice among several state-of-the-art methods, including eight publicly released pretrained medical models, at every label budget. An MAE pretrained on the same images with the same encoders and optimization settings trails GALA-3D by 2.3 to 5.8 Dice points from one labelled case to 10\% of the labels, which attributes the gain over masked reconstruction to the pretraining framework rather than to the data. Pretrained on a head CT cohort instead, GALA-3D becomes the best method on a haemorrhage segmentation benchmark whenever labels are limited, which suggests that the framework transfers to a new anatomical region.
    \end{itemize}

    \section{Related Work}
    \label{sec:related}

    \subsection{Masked image modelling and latent prediction}

    Early medical SSL methods learned from image-space pretext tasks such as restoration, spatial transformation prediction, context prediction, and inpainting \citep{modelgenesis,rubiks,swinunetr}. MAEs \citep{he_masked_2021} turned inpainting into a scalable objective by reconstructing a large masked portion of the input from the visible remainder. MIM has since been adapted to 3D medical imaging to address the challenges of volumetric data and multi-scale anatomical representation learning, with convolutional, transformer, and hybrid architectures \citep{medicalmim,spark3d,hyspark,hi_end_mae}. To place more emphasis on the encoder, Hi-End-MAE \citep{hi_end_mae} introduces encoder-driven reconstruction and hierarchical decoding, and HySparK \citep{hyspark} exploits hierarchical representations during masked pretraining. However, these methods still reconstruct voxel intensities, a target that does not explicitly organize the representation space around the semantic properties that downstream tasks require.

    Latent prediction addresses this limitation by replacing the image-space target with a representation-space one. iBOT \citep{zhou_ibot_2022} supervises masked patch predictions with the representations of an unmasked teacher, and I-JEPA \citep{assran_self-supervised_2023} predicts the latent representations of target blocks from visible context. DINOv2 \citep{dinov2} and its 3D medical adaptation, 3DINO \citep{3dino}, add image-level self-distillation to masked patch prediction. All of these methods draw their targets from a momentum-updated teacher, a second copy of the encoder with its own update schedule and computational overhead. In contrast, GALA-3D draws its masked-patch targets from a clean pass of the same encoder, so no separate teacher is maintained.

    \begin{figure*}
        \centering
        \includegraphics[width=0.97\linewidth]{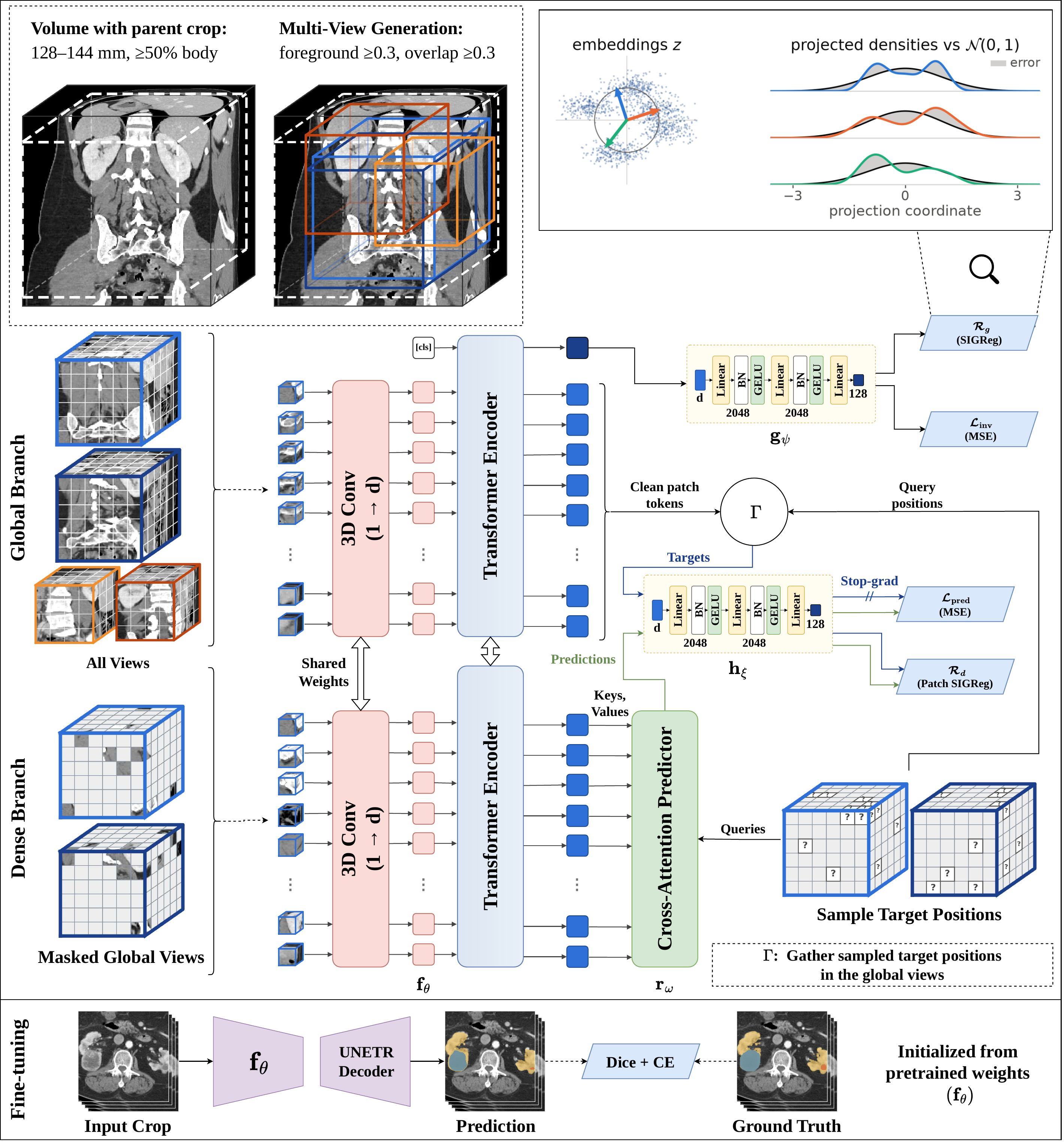}
        \caption{\textbf{Overview of GALA-3D.} Global (blue) and local (orange) views from a shared foreground-anchored parent region (top left) are encoded by one weight-shared 3D ViT $f_\theta$. \emph{Global branch:} projected class tokens are pulled towards the mean of the global views and regularized by Sketched Isotropic Gaussian Regularization (SIGReg) \citep{lejepa}. \emph{Dense branch:} from the visible tokens of a masked global view, a cross-attention predictor regresses the clean patch tokens of the unmasked view at queried positions ($\Gamma$, stop-gradient), with a patch-level SIGReg. \emph{Fine-tuning:} only $f_\theta$ is kept and trained with a UNETR decoder \citep{unetr}. [cls]: class token. MSE: mean squared error. Dice + CE: Dice and cross-entropy losses. Values follow our GALA configuration (Table~\ref{tab:gala-hparams}).}
        \label{fig:method}
    \end{figure*}

    \subsection{Contrastive and multi-view learning}

    Contrastive methods such as SimCLR \citep{chen_simple_2020} and MoCo \citep{moco} encourage agreement between augmented views of one sample while separating them from the views of other samples, and similar objectives have been adapted to volumetric images using relationships between images, crops, anatomical regions, or spatial locations \citep{voco,vox2vec}. In medical imaging, different patients contain similar anatomy and can form false negatives, while distant regions of one large volume can depict different anatomy and need not form a meaningful positive pair \citep{boosting_contrastive_learning,anatomy_aware_contrastive}. Some methods therefore build anatomical or spatial information into the objective. Anatomy-aware contrastive learning \citep{anatomy_aware_contrastive} defines positive and negative pairs from anatomy-category labels, and VoCo \citep{voco} predicts the position of a crop relative to the regions of its volume. However, both still push some representations apart, anatomy-aware contrastive learning through negative pairs and VoCo by enforcing dissimilarity between different regions of the same volume, on the assumption that they contain different organs. In contrast, GALA-3D uses anatomy to constrain view sampling, drawing every global and local view from a shared foreground-anchored parent region so that all positive views depict related anatomy, without negative pairs.

    \subsection{Non-contrastive joint-embedding methods}

    Non-contrastive methods such as BYOL \citep{grill2020bootstrap}, SimSiam \citep{simsiam}, and DINO \citep{Caron2021EmergingTransformers} learn from positive views alone and prevent collapse with predictor asymmetry, momentum-updated teachers, and other engineering heuristics. DINOv2 \citep{dinov2} and its medical extensions, 3DINO \citep{3dino} and TAP-CT \citep{tapct}, show the value of combining global view consistency with local representation learning, but adapting them to 3D required dedicated changes to the training setup \citep{3dino,tapct}. LeJEPA \citep{lejepa} instead prevents collapse with SIGReg, which regularizes the embeddings towards an isotropic Gaussian distribution without complex engineering heuristics. DeJEPA \citep{dejepa} applied SIGReg to dense voxel embeddings matched between two overlapping crops, pooling the embeddings of all sampled voxels. GALA-3D applies SIGReg at both the image and the patch level, so that global consistency and dense latent prediction are learned in a single encoder without a teacher network.

    \subsection{Medical imaging foundation models}

    Several of these objectives have since been scaled into publicly released foundation models, mainly using contrastive learning and self-distillation. Contrastive learning underlies VoCo \citep{voco}, whose contextual position prediction was extended to the 160,000 public CT volumes of PreCT-160K \citep{voco_v2}, and CT-FM \citep{ctfm}, pretrained on 148,000 CT images from 69 Imaging Data Commons (IDC) cohorts. Self-distillation has followed a similar path. 3DINO \citep{3dino} adapts DINOv2 to 3D images and is pretrained on about 100,000 CT, MRI, and positron emission tomography (PET) volumes, and Curia-2 \citep{curia2} further tailors the augmentations and objective functions of DINOv2 to radiological data, using clinical CT and MRI from a single hospital with a 2D slice-based encoder. By comparison, GALA-3D is pretrained on the MERLIN images \citep{merlin}, a cohort approximately six times smaller than PreCT-160K and the CT-FM corpus and, unlike those used by some compared methods, entirely disjoint from all evaluation benchmarks.

    \section{Methodology}
    \label{sec:methodology}

    GALA-3D learns complementary volume- and patch-level representations from unlabelled images. As illustrated in Figure~\ref{fig:method}, it combines an anatomy-aware multi-view sampler and a weight-shared 3D Vision Transformer (ViT), described in Section~\ref{sec:global-local-representation-learning}, and a global and a dense representation-space loss, described in Section~\ref{sec:pretraining-objective}. A clean pass and a masked, visible-token pass through the same encoder provide the target and context representations, so that no momentum encoder is required. Sections~\ref{sec:implementation-details} and~\ref{sec:downstream-adaptation} then give the pretraining setup and the protocol used to transfer the encoder to segmentation.

    \subsection{Anatomy-aware views and shared encoder}
    \label{sec:global-local-representation-learning}

    \paragraph{Anatomy-aware multi-view sampling} Let $x_b$, $b=1,\ldots,B$, denote the $b$-th of the $B$ distinct images in a mini-batch, each accompanied by a foreground body mask. Independently sampling crops from a large volume does not guarantee a meaningful positive pair because the crops may contain disjoint anatomical regions. The sampler is therefore anatomy-aware through its parent construction, as every view of an image is drawn from one shared parent region placed on the body, so that all positive views depict related anatomy at different fields of view. To build this parent, we draw a foreground anchor $a_b$ from the body mask and use it to place a cubic parent region $P_b \subset x_b$ of randomly sampled edge length that lies entirely inside the volume. Candidate parents are scored by their foreground occupancy. The first candidate whose occupancy reaches $\tau_P$ is accepted, and otherwise the highest-occupancy candidate is retained.

    For each image, we draw $V_g$ global views and $V_l$ local views from the parent region. Global views cover a larger field of view than local views, loosely mirroring how radiologists read an exam, first surveying it globally and then zooming in on specific regions. Each global crop $x^{g}_{b,v}$ must contain $a_b$ and have foreground occupancy of at least $\tau_F$. The $m$-th local view, $m=1,\ldots,V_l$, is paired with global view $1+\left((m-1)\bmod V_g\right)$. It must have foreground occupancy of at least $\tau_F$ and have at least a fraction $\tau_O$ of its volume inside the paired global crop. This produces the set $\mathcal{X}_b=\{x^{g}_{b,v}\}_{v=1}^{V_g}\cup\{x^{l}_{b,m}\}_{m=1}^{V_l}$ of $V=V_g+V_l$ views, whose members vary in position and field of view while remaining anatomically related by construction. We index the members of $\mathcal{X}_b$ by $v=1,\ldots,V$, with the $V_g$ global views first, and write $x_{b,v}$ for a view of either type. Each view type is resampled to its own fixed input grid, which turns the random field of view into a scale augmentation and lets views of one type be batched. Global and local views are therefore encoded at their own resolutions, rather than resampling the local crops to the global input grid.

    Spatial and intensity transformations are sampled independently for every view. Besides rotations and flips, these include a randomized clinical intensity window, so that the encoder observes the same anatomy under soft-tissue, lung, and bone contrast, mirroring standard window settings in clinical practice. Moreover, windowing stretches the tissues of interest across the input range, which a linear scaling of the full HU range would confine to a narrow band.

    \paragraph{Weight-shared multi-resolution encoder} A single 3D ViT \citep{dosovitskiy_image_2021} $f_{\theta}$ encodes every global and local view. For a view $x_{b,v} \in \mathbb{R}^{1\times D_v\times H_v\times W_v}$, where $D_v\times H_v\times W_v$ is the size of its input grid, and a cubic patch edge $p$, a strided 3D convolution produces $N_v$ patch tokens of dimension $d$. A learnable class token is added to the sequence. Rather than using a learned absolute position table, GALA-3D applies 3D rotary position embeddings (RoPE) \citep{rope} to centred patch-grid coordinates. The same encoder can consequently process global and local token grids without interpolating positional embeddings. A clean forward pass, on the full view without masking, gives $(c_{b,v},H_{b,v})=f_{\theta}(x_{b,v})$, where $c_{b,v}\in\mathbb{R}^{d}$ is the output class token and $H_{b,v}\in\mathbb{R}^{N_v\times d}$ contains the spatial patch representations. We denote by $N_g$ and $N_l$ the number of patch tokens of a global and a local view, respectively. The encoder parameters $\theta$ are shared across views, resolutions, and the clean and masked branches described below.

    \begin{figure*}[!tp]
        \centering
        \includegraphics[width=0.975\linewidth]{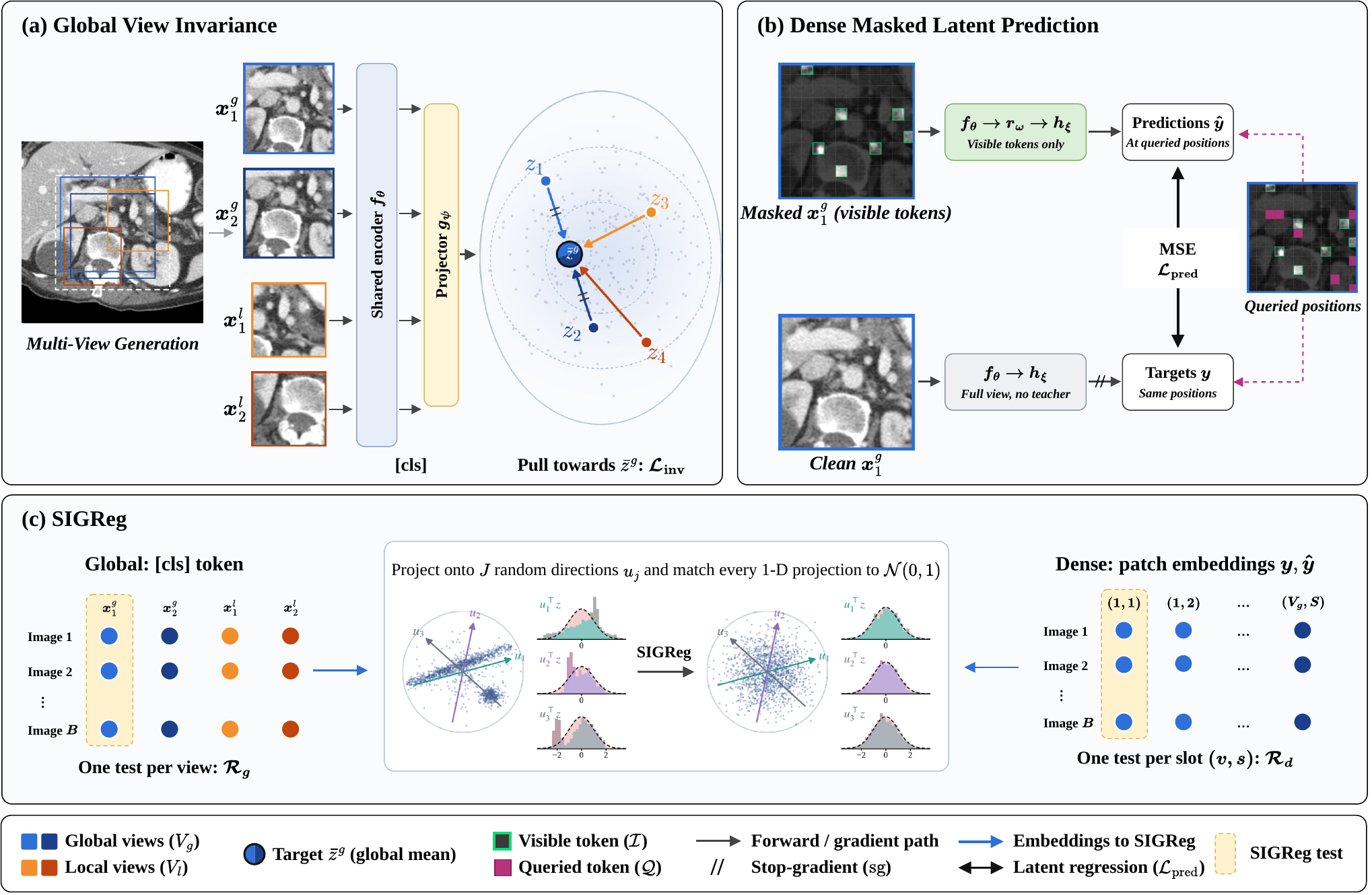}
        \caption{\textbf{Conceptual view of the GALA-3D losses.} \textbf{(a)} Global view invariance: for one image $x_b$, the global ($x^{g}_{1}, x^{g}_{2}$) and local ($x^{l}_{1}, x^{l}_{2}$) views drawn from one anatomy-aware parent region are encoded by the shared encoder $f_\theta$ and projected by $g_\psi$, and every projected class token is pulled towards $\bar z^{g}$, the mean of the global-view embeddings ($\mathcal{L}_{\mathrm{inv}}$, Eq.~\eqref{eq:global-invariance}). \textbf{(b)} Dense masked latent prediction: from the visible tokens of the masked global view $x^{g}_{1}$, $f_\theta$, the predictor $r_\omega$ and the patch head $h_\xi$ predict the representations $\hat y$ of queried masked positions, which are regressed onto stop-gradient targets $y$ from a clean pass of the same encoder on the full view, without a teacher network ($\mathcal{L}_{\mathrm{pred}}$, Eq.~\eqref{eq:dense-prediction}). \textbf{(c)} SIGReg projects embeddings onto $J$ random directions $u_j$ and matches each one-dimensional projection to $\mathcal{N}(0,1)$. The images of the batch form the sample axis: the global term $\mathcal{R}_g$ (Eq.~\eqref{eq:global-sigreg}) runs one test per view, and the dense term $\mathcal{R}_d$ (Eq.~\eqref{eq:dense-sigreg}) one test per slot $(v,s)$, on both targets and predictions. [cls]: class token.}
        \label{fig:objectives}
    \end{figure*}

    \subsection{GALA-3D pretraining losses}
    \label{sec:pretraining-objective}

    \paragraph{Global view-consistency objective} GALA-3D combines a global and a dense objective (Figure~\ref{fig:objectives}). The class token $c_{b,v}$ of each view is mapped by a multilayer projection head $g_{\psi}:\mathbb{R}^{d}\rightarrow\mathbb{R}^{k}$ with parameters $\psi$ to an embedding $z_{b,v}=g_{\psi}(c_{b,v})$ of dimension $k$. Since the global views are indexed first, $z^{g}_{b,v}=z_{b,v}$ with $v\le V_g$ denotes the embedding of the $v$-th global view of image $b$, and $\bar z^{g}_b=\frac{1}{V_g}\sum_{v=1}^{V_g}z^{g}_{b,v}$ denotes the mean over its global views. Only the global views, which cover the larger field of view, form the target mean, while both global and local views are predictions:
    
    \begin{equation}
        \mathcal{L}_{\mathrm{inv}}
        = \frac{1}{B V k}
        \sum_{b=1}^{B}\sum_{v=1}^{V}
        \left\|z_{b,v}-\bar z^{g}_b\right\|_2^2 .
        \label{eq:global-invariance}
    \end{equation}

    The invariance term alone admits a collapsed solution. SIGReg, introduced as part of LeJEPA, addresses this degeneracy by matching the distribution of projected representations to an isotropic Gaussian. Following its original formulation, let $\mathcal{Z}$ be a set of $B$ embeddings, one from each independent image of the batch, and let $u_j\in\mathbb{R}^{k}$ be a random unit direction. The empirical characteristic function of the projections, evaluated at $t\geq 0$, is:
    \begin{equation}
        \widehat{\varphi}_{\mathcal{Z},j}(t)
        = \frac{1}{B}\sum_{z\in\mathcal{Z}}
        \exp\!\left(\mathrm{i}\,t\,u_j^{\top}z\right),
    \end{equation}
    and SIGReg compares it with the characteristic function of a standard Gaussian, $\varphi_0(t)=\exp(-t^2/2)$:
    \begin{equation}
        \operatorname{SIGReg}(\mathcal{Z})
        = \frac{2B}{J}\sum_{j=1}^{J}
        \int_{0}^{T}
        w(t)\left|
        \widehat{\varphi}_{\mathcal{Z},j}(t)-\varphi_0(t)
        \right|^2\,\mathrm{d}t,
        \label{eq:sigreg}
    \end{equation}
    where $w(t)=\exp(-t^2/2)$ is a Gaussian weighting, as in the reference implementation of LeJEPA, $T$ bounds the integration range, and $J$ unit directions are freshly sampled at each evaluation. The factor $B$ is the sample-size scaling of the Epps--Pulley statistic \citep{epps-pulley}, the characteristic-function test of normality on which SIGReg is built. The global regularizer averages this statistic over the views:
    \begin{equation}
        \mathcal{R}_{g}
        = \frac{1}{V}\sum_{v=1}^{V}\operatorname{SIGReg}(\mathcal{Z}_v),
        \qquad
        \mathcal{Z}_v=\{z_{b,v}\}_{b=1}^{B},
        \label{eq:global-sigreg}
    \end{equation}
    so that each view distribution is estimated across the $B$ distinct images of the batch. Different crops from the same image remain on the view axis and are not treated as independent samples. A weight $\lambda_g\in[0,1]$ balances view invariance against this distributional regularization in the overall objective (Eq.~\eqref{eq:overall-loss}). We write $\mathcal{L}$ for the prediction terms and $\mathcal{R}$ for the SIGReg regularizers. The view-invariance and SIGReg terms are inherited from LeJEPA. Our contribution at the global level lies in constructing anatomically valid positive views for volumetric images. The dense objective below extends this framework to leverage patch tokens, with a patch-level SIGReg that prevents the collapse of teacher-free dense prediction.

    \paragraph{Dense masked latent prediction} The global objective summarizes a complete crop in a single token and does not directly require spatially resolved representations. Therefore, we add a dense prediction task on the patch tokens of the $V_g$ global views. Local views contribute only to the global terms and therefore incur no masked encoder or predictor pass. For each global view, a masked index set $\mathcal{M}_{b,v}$ containing a fraction $\rho$ of its $N_g$ patches is sampled uniformly, and $\mathcal{I}_{b,v}=\{1,\ldots,N_g\}\setminus\mathcal{M}_{b,v}$ denotes the visible indices. As in MAEs, the masked tokens are discarded immediately after patch embedding, so that the transformer blocks process only the visible tokens. Without the output of the class token, the resulting patch outputs form the visible-context representation $\widetilde H_{b,v}=f_{\theta}(x_{b,v};\mathcal{I}_{b,v})_{\mathrm{patch}}\in\mathbb{R}^{|\mathcal{I}_{b,v}|\times d}$.
    The clean patch tokens $H_{b,v}$ from the full-token pass provide online latent targets.

    Predicting every masked location is unnecessary at a high masking ratio \citep{crossmae}. For each view of each image, we sample a query set $\mathcal{Q}_{b,v}\subseteq\mathcal{M}_{b,v}$ containing a fraction $q$ of the masked positions. The full mask is retained for the visible encoder, and query subsampling reduces only the predictor and target-projection work. A predictor $r_{\omega}$ with parameters $\omega$ uses learnable mask-token queries at the coordinates of $\mathcal{Q}_{b,v}$ and cross-attends to the visible tokens at the coordinates of $\mathcal{I}_{b,v}$, giving $\widehat H_{b,v}^{\mathcal{Q}}=r_{\omega}(\widetilde H_{b,v},R_{b,v}^{\mathcal{I}},R_{b,v}^{\mathcal{Q}})$, where $R^{\mathcal{I}}_{b,v}$ and $R^{\mathcal{Q}}_{b,v}$ denote the centred patch-grid coordinates of the visible and queried patches, which condition the RoPE attention \citep{rope}.
    Following CrossMAE \citep{crossmae}, the predictor is a stack of cross-attention and feed-forward blocks. Its attention is conditioned by RoPE, and its width and number of attention heads match those of the encoder. It contains no self-attention among the mask queries: every prediction must therefore be inferred directly from visible encoder context.

    A second projection head $h_{\xi}:\mathbb{R}^{d}\rightarrow\mathbb{R}^{k}$ with parameters $\xi$ and the same architecture as $g_{\psi}$ projects both predictions and clean targets. Writing $\hat y_{b,v,i}=h_{\xi}(\widehat H_{b,v,i})$ and $y_{b,v,i}=h_{\xi}(H_{b,v,i})$ for the projected prediction and clean target at patch position $i\in\mathcal{Q}_{b,v}$, the dense prediction loss is
    \begin{equation}
        \mathcal{L}_{\mathrm{pred}}
        = \frac{1}{k\,N_{\mathcal{Q}}}
        \sum_{b,v}\sum_{i\in\mathcal{Q}_{b,v}}
        \bigl\|\hat y_{b,v,i}-\operatorname{sg}[y_{b,v,i}]\bigr\|_2^2,
        \label{eq:dense-prediction}
    \end{equation}
    where $N_{\mathcal{Q}}=\sum_{b,v}|\mathcal{Q}_{b,v}|$ is the total number of queried positions and $\operatorname{sg}$ denotes stop-gradient \citep{simsiam}. Stop-gradient is used only for the attraction target in Eq.~\eqref{eq:dense-prediction}. The clean branch is still optimized by the global loss and the dense distributional regularizer. Because targets are produced online by the same encoder and patch projector, the method introduces no separate target-network parameters.

    To prevent the patch representations from collapsing, SIGReg is also applied to the projected clean targets and predictions. As in the global branch, the regularizer must be computed across independent images, so patches of one image are not pooled. Instead, for each global view, $S$ queried positions are drawn per image, independently for targets and predictions, which forms $V_gS$ slots that each contain one embedding from every image of the batch. SIGReg is evaluated on each slot, with the $B$ images as its sample axis, and the results are averaged:
    \begin{equation}
        \mathcal{R}_{d}
        = \frac{1}{2V_gS}\sum_{v=1}^{V_g}\sum_{s=1}^{S}
        \Big[\operatorname{SIGReg}\bigl(\mathcal{Y}_{v,s}\bigr)
        + \operatorname{SIGReg}\bigl(\widehat{\mathcal{Y}}_{v,s}\bigr)\Big],
        \label{eq:dense-sigreg}
    \end{equation}
    where $\mathcal{Y}_{v,s}=\{y_{b,v,i_{b,v,s}}\}_{b=1}^{B}$ and $\widehat{\mathcal{Y}}_{v,s}=\{\hat y_{b,v,i'_{b,v,s}}\}_{b=1}^{B}$ are the target and prediction slots, whose positions $i_{b,v,s}$ and $i'_{b,v,s}$ are drawn from $\mathcal{Q}_{b,v}$. A weight $\lambda_d\in[0,1]$ balances dense prediction against this regularizer.

    Slot-wise SIGReg also mitigates the positional collapse observed in teacher-free dense prediction (Section~\ref{sec:ablations}). A representation dominated by patch position varies along only a few directions. Because each slot pools embeddings from different images at random positions, such a representation concentrates the variance of the slot in few directions of the projection space, which the random one-dimensional projections of SIGReg detect as a departure from isotropy. SIGReg therefore penalizes the low-rank positional codes that we observe when the patch-level regularizer is removed.

    \paragraph{Overall objective} The final self-supervised objective sums the global and dense terms, each balanced by its own weight:
    \begin{equation}
        \mathcal{L}_{\mathrm{GALA\text{-}3D}}
        = (1-\lambda_g)\mathcal{L}_{\mathrm{inv}} + \lambda_g\mathcal{R}_{g}
        + (1-\lambda_d)\mathcal{L}_{\mathrm{pred}} + \lambda_d\mathcal{R}_{d}.
        \label{eq:overall-loss}
    \end{equation}
    The clean global encoder pass is shared by the global representation and dense target branches. Relative to this pass, dense training adds, for each global view, an encoder sequence of $|\mathcal I|\approx(1-\rho)N_g$ visible patch tokens and $|\mathcal Q|\approx q\rho N_g$ predictor queries. Because attention cost grows quadratically with sequence length, high-ratio masking and query subsampling make dense supervision practical.

    \subsection{Pretraining setup}
    \label{sec:implementation-details}

    GALA-3D is pretrained on the MERLIN cohort (Section~\ref{sec:datasets}) with AdamW \citep{adamw} on batches of 64 distinct images for 100,000 updates in mixed bfloat16 precision, with base learning rates of $5\times10^{-4}$ for ViT-S/8 and $2.5\times10^{-4}$ for ViT-B/8, and the two resulting models are denoted GALA-S and GALA-B. Each encoder is pretrained on a single NVIDIA A100 40\,GB GPU, in 30 hours for ViT-S/8 and 60 hours for ViT-B/8. Encoder sizes are listed in Table~\ref{tab:model-capacity}. All pretraining settings (Table~\ref{tab:gala-hparams}) and the MAE3D control are detailed in \ref{app:pretraining}.

    \begin{table}[!tp]
        \caption{\textbf{Encoder configurations.} Parameter counts are for the encoder only, and LR is the GALA-3D pretraining learning rate. GALA-S, GALA-B: GALA-3D pretraining with the ViT-S/8 and ViT-B/8 encoders. UNETR-S, UNETR-B: the same encoders with a UNETR decoder, trained from scratch.}
        \label{tab:model-capacity}
        \vspace{5pt}
        \centering
        \small
        \begin{adjustbox}{max width=\columnwidth}
            % Hand-written table, moved out of main.tex on 2026-11-17. Edit here.
\begin{tabular}{lcc}
    \toprule
    & ViT-S/8 & ViT-B/8 \\
    \midrule
    Width & 384 & 768 \\
    Heads & 6 & 12 \\
    Blocks & 12 & 12 \\
    Parameters & 21M & 85M \\
    LR & $5\times10^{-4}$ & $2.5\times10^{-4}$ \\
    \bottomrule
\end{tabular}

        \end{adjustbox}
    \end{table}

    \subsection{Downstream adaptation}
    \label{sec:downstream-adaptation}

    All downstream tasks are 3D medical image segmentation (Section~\ref{sec:datasets}). The GALA transfer protocol fine-tunes the complete encoder with a newly initialized UNETR decoder \citep{unetr} that reads intermediate encoder blocks. Volumes are resampled to $1\,\mathrm{mm}$ isotropic spacing, and CT intensities are clipped to a clinical window, $[-175,250]$ Hounsfield units (HU) for abdominal CT and $[0,100]$\,HU for head CT, and rescaled. ISLES22 volumes are resampled in the same way and z-normalized per channel within the head foreground. Model-specific settings of each external model, such as the encoder blocks read by the decoder for different encoder sizes and preprocessing, follow the specifications of each architecture.

    Training uses AdamW with a cosine schedule on $96^3$ crops balanced between classes, or between foreground and background for nested targets, with an update budget that grows with the label budget, early stopping on validation Dice, and sliding-window testing of the best validation checkpoint. Datasets whose targets are nested, such as an organ and the lesions it contains, are formulated as region-based multi-label segmentation, and the others as multi-class segmentation (\ref{app:finetuning}). A learning rate sweep was done on a subset of low-label KiTS23 \citep{kits23} validation folds without using test data, and each selected learning rate was then applied to every dataset and label budget. The complete fine-tuning settings are given in \ref{app:finetuning}.

    \section{Experiments}
    \label{sec:experiments}

    \subsection{Datasets}
    \label{sec:datasets}

    \paragraph{Pretraining cohort} We pretrain GALA-3D and the matched MAE baselines (Section~\ref{sec:comparative-methods}) on the MERLIN cohort \citep{merlin} of abdominal CT, resampled to a $1\,\mathrm{mm}$ isotropic grid. The release lists 25,494 images, which correspond to 25,489 distinct images. After excluding 75 images that could not be preprocessed or could not accommodate a training view (Figure~\ref{fig:cohort}), 25,414 images form the pretraining set. No diagnosis, report, or segmentation labels are used during self-supervised pretraining.

    \paragraph{Downstream datasets} Transfer capabilities to the downstream tasks are evaluated on eight segmentation datasets that probe three degrees of distance from the pretraining domain (Table \ref{tab:downstream-datasets}). Five are abdominal CT and therefore in-domain for MERLIN: KiTS23 \citep{kits23,kits19}, three tasks of the Medical Segmentation Decathlon (MSD) \citep{Antonelli2022-qy} (hepatic vessel, pancreas, and colon), and PANORAMA \citep{panorama}, whose targets range from whole organs to thin vessels and small, infrequent lesions. Two are non-contrast head CT, which keep the modality but change the anatomy: MSICH \citep{ichseg}, which annotates images from the RSNA 2019 haemorrhage challenge \citep{flanders_construction_2020}, and INSTANCE22 \citep{li_state---art_2023,hematoma_jbhi}. The last dataset, ISLES22 \citep{isles22}, is a brain MRI dataset with diffusion-weighted and apparent diffusion coefficient images. It changes both anatomy and modality and is analysed separately (Section~\ref{sec:mri}).

     Within each of these three groups, we selected public datasets that target lesions and fine structures, such as tumours, vessels, and haemorrhages. These are challenging and costly to annotate, which is where label-efficient pretraining is most needed. KiTS23, MSD Pancreas, and PANORAMA also segment the organ that contains the lesion, the kidney or the pancreas, so that organ segmentation, the target of common multi-organ benchmarks, is evaluated alongside lesion segmentation. Benchmarks on which performance is saturated or on which statistical noise can exceed differences between methods discriminate poorly between segmentation methods, whereas KiTS23, our largest dataset, is among the benchmarks recommended for comparing them \citep{isensee2024nnunetrevisited}. From PANORAMA we use the 479 cases with manual lesion annotations, 97 of which are MSD pancreas images that PANORAMA annotated independently (median tumour Dice between the two label sets: 0.77). The same images therefore appear, with different labels, in our MSD Pancreas set, but each dataset is cross-validated and fine-tuned separately, so this sharing does not place any test image in training.

    \begin{table}[!tp]
        \caption{\textbf{Downstream segmentation datasets.} Cases are the labelled cases used for five-fold cross-validation, in which each fold holds out about 20\% for testing and 10\% for validation, and the remaining 70\% form the training pool. Mod.: Modality, Ch.: input channels, PDAC: pancreatic ductal adenocarcinoma, ICH: intracerebral haemorrhage, IVH: intraventricular haemorrhage, PHE: perihaematomal oedema.}
        \label{tab:downstream-datasets}
        \vspace{5pt}
        \centering
        \footnotesize
        \setlength{\tabcolsep}{2.5pt}
        \begin{adjustbox}{max width=\columnwidth}
            % Hand-written table, moved out of main.tex on 2026-11-17. Edit here.
\begin{tabular}{lllcc}
    \toprule
    Dataset & Mod. & Targets & Ch. & Cases \\
    \midrule
    \multicolumn{5}{l}{\emph{Abdominal CT (in-domain)}} \\
    KiTS23 & CT & kidney, masses, tumour & 1 & 489 \\
    MSD H.\,vessel & CT & vessel, tumour & 1 & 303 \\
    MSD Pancreas & CT & pancreas, tumour & 1 & 281 \\
    PANORAMA & CT & pancreas, PDAC & 1 & 479 \\
    MSD Colon & CT & colon tumour & 1 & 126 \\
    \multicolumn{5}{l}{\emph{Head CT (same modality, out of domain)}} \\
    MSICH & CT & ICH, IVH, PHE & 1 & 359 \\
    INSTANCE22 & CT & haemorrhage & 1 & 100 \\
    \multicolumn{5}{l}{\emph{Brain MRI (different modality)}} \\
    ISLES22 & MRI & infarct & 2 & 250 \\
    \bottomrule
\end{tabular}

        \end{adjustbox}
    \end{table}

    \paragraph{Cross-validation and labelled subsets} Every dataset is split into five folds which are disjoint and together cover every labelled case. Within each fold, a fixed validation partition is used for checkpoint selection and the remaining cases form the training pool. The partitions are identical for every method.

    Each fold is trained under four label budgets: one labelled case, 1\%, 10\%, and 100\% of the training pool. Fractional subset sizes are rounded to the nearest whole case with a minimum of one, and the 1\% subset is contained in the 10\% subset, which is contained in the full pool. The 1\% budget is reported only where it is distinct from one labelled case (KiTS23, MSD hepatic vessel, MSD Pancreas, PANORAMA, MSICH, and ISLES22), which gives 26 dataset--budget settings on CT, 19 of them with one labelled case to 10\% of the labels, and four on MRI. Each reported value is the mean over the five folds.

    \subsection{Comparative methods}
    \label{sec:comparative-methods}

    GALA-3D is compared with eight external pretrained models and with networks trained from random initialization. External references use each method's released weights and native encoder: Models Genesis \citep{modelgenesis}, the self-supervised Swin UNETR of \citet{swinunetr}, denoted Swin-SSL, VoCo-B and VoCo-L \citep{voco_v2}, Hi-End-MAE \citep{hi_end_mae}, 3DINO \citep{3dino}, CT-FM \citep{ctfm}, and Curia-2 \citep{curia2}. ViT encoders are fine-tuned with a UNETR decoder \citep{unetr} and Swin encoders with Swin UNETR \citep{hatamizadeh_swin_2022}, the segmentation networks with which several volumetric foundation models have been evaluated \citep{swinunetr,voco,hi_end_mae}, while Models Genesis and CT-FM keep their native 3D U-Net \citep{ronneberger_u-net_2015} and SegResNet \citep{segresnet}, and the 2D Curia-2 encoder is applied slice by slice. UNETR with each encoder (UNETR-S, UNETR-B) and Swin UNETR, trained from scratch, serve as baselines. Self-configuring pipelines such as nnU-Net \citep{Isensee2021-qk} change the architecture, preprocessing, and training recipe together, which would confound the contribution of pretraining, and are therefore not included.

    Masked voxel reconstruction with MAE3D \citep{he_masked_2021} serves as a same-data control. Its variants, MAE-S and MAE-B, share the encoders, pretraining cohort, preprocessing, optimization settings, update budget, and transfer protocol of GALA-3D, so that the difference between the two isolates the pretraining framework. Because VoCo and Hi-End-MAE already report improvements over MAE, MAE3D is treated as an ablation of the framework rather than as a competing method and is reported in Section~\ref{sec:objective-control}.

    \paragraph{Prior exposure to the evaluation data} Table~\ref{tab:overlap} (\ref{app:overlap}) summarizes the pretraining data of every compared model: several external models were pretrained on corpora that contain some of our public evaluation datasets, notably KiTS (VoCo, Hi-End-MAE, 3DINO, and CT-FM), the MSD tasks and the PANORAMA cases derived from them (VoCo and Hi-End-MAE), the rest of PANORAMA (VoCo), and part of the MSICH images (3DINO), and several were pretrained with head or head-and-neck CT. In comparison, GALA-3D and MAE3D are pretrained on MERLIN alone, which contains no head images and is disjoint from every evaluation dataset. The sources of each entry are detailed in \ref{app:overlap}.

    \subsection{Evaluation and statistical analysis}
    \label{sec:evaluation}

    We evaluate the mean foreground Dice similarity coefficient (DSC), computed per case, averaged over the test fold and then over the five folds, with equal weights for each foreground class within a dataset and for each dataset across datasets. Boundary agreement is measured with the normalized surface distance (NSD) \citep{nikolov_nsd} at a tolerance of $2\,\mathrm{mm}$, or two voxels at the working resolution.

    As all methods share the same partitions, each (dataset, fold) pair gives one paired observation per label budget. Each GALA-3D encoder is compared with every other method except MAE3D, which is tested separately (Table~\ref{tab:matched}), by two-sided Wilcoxon signed-rank tests \citep{wilcoxon} on these fold-level DSC values over the seven CT datasets, with Holm correction \citep{holm} across comparators and budgets. Throughout, a difference is called significant when its Holm-adjusted p-value is below 0.05. Confidence intervals are 95\% bootstrap intervals \citep{bootstrap} over the same pairs. Because the folds of one dataset are not fully independent, we also use a two-level bootstrap that first resamples datasets and then folds within each selected dataset and gives equal weight per dataset.

    \begin{figure*}[!t]
        \centering
        \includegraphics[width=\textwidth]{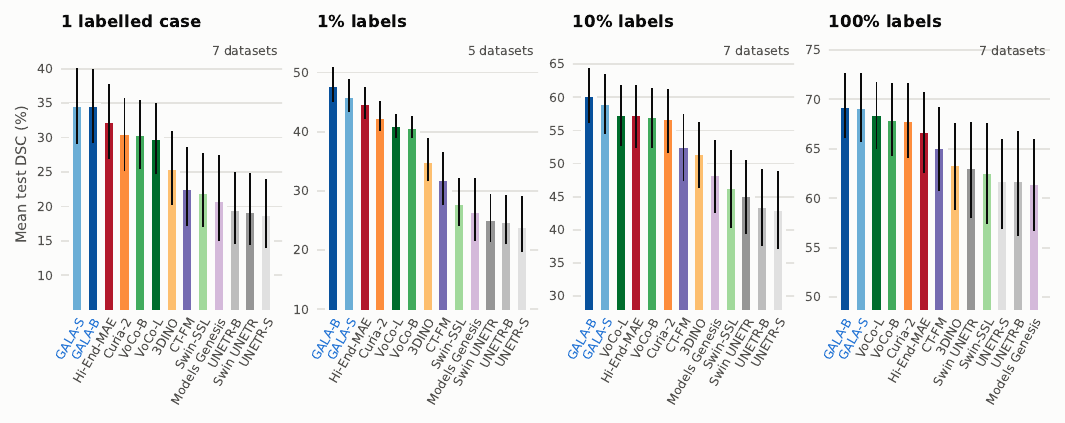}
        \caption{\textbf{Mean held-out-test DSC (\%) of every method at each label budget on the seven CT datasets (five at 1\%).} Bars are sorted, and error bars are 95\% bootstrap confidence intervals over (dataset, fold) pairs. Each method keeps its colour in every panel. Blue: GALA-3D. Green: Swin-based external models. Red and orange: ViT-based external models. Purple: convolutional external models. Grey: training from scratch.}
        \label{fig:budget-bars}
    \end{figure*}
    \begin{figure*}[!t]
        \centering
        \includegraphics[width=\textwidth]{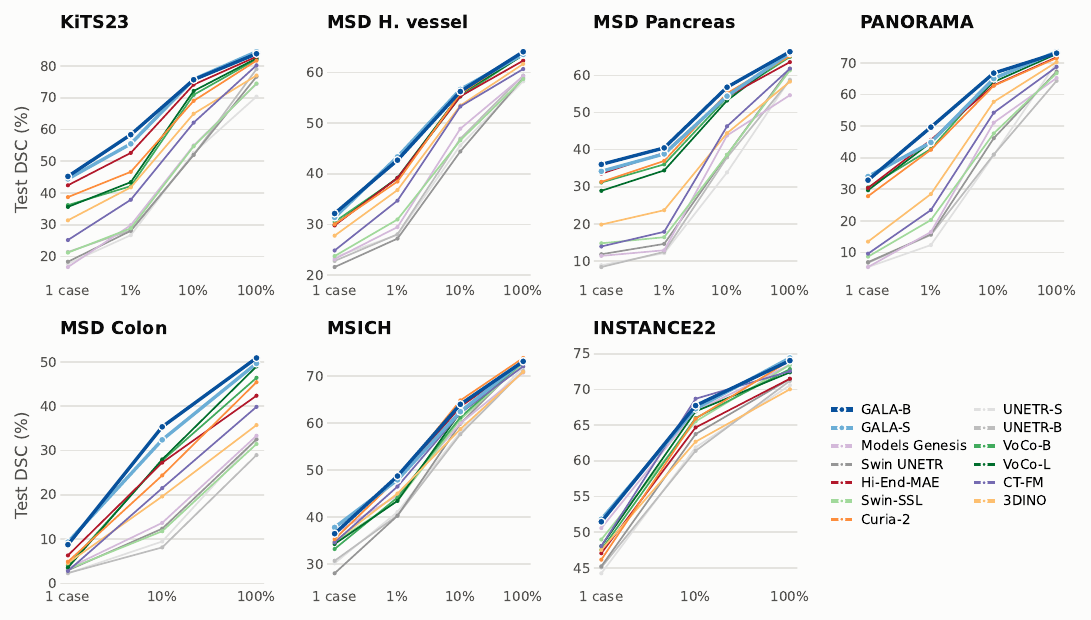}
        \caption{\textbf{Held-out-test DSC (\%) against label budget on each CT dataset.} Colours as in Figure~\ref{fig:budget-bars}, with the GALA-3D encoders drawn thicker. The benefit of pretraining is largest at the lowest label budgets, and on the abdominal datasets GALA-3D is at or above every other method at every budget.}
        \label{fig:label-efficiency}
    \end{figure*}
    \begin{table*}[!t]
        \caption{\textbf{Held-out-test DSC (\%) of every method on the CT datasets with one labelled case and 1\% of the labels.} Mean $\pm$ standard deviation over five folds. Mean: average over the datasets. Bold marks the best method in each column and underline the second best. In (b), datasets on which 1\% equals one labelled case (MSD Colon, INSTANCE22) are omitted. $^{\dagger}$GALA-S pretrained on RSNA 2019 head CT with MSICH patients removed (\ref{app:headct}), shown for reference on the head-CT datasets and not ranked. $^{\blacktriangle}$Higher than every other method in the column.}
        \label{tab:res-low}
        \centering
        \footnotesize
        \setlength{\tabcolsep}{3pt}
        \captionsetup[subtable]{skip=2pt}
        \begin{subtable}{\textwidth}
            \caption{One labelled case}
            \label{tab:res-one_shot}
            % Generated by downstream/NAB/plot/make_tex_tables.py; do not edit by hand.
\begin{tabular*}{\textwidth}{@{\extracolsep{\fill}}lr*{7}{c}c}
  \toprule
   & & \multicolumn{5}{c}{Abdominal CT} & \multicolumn{2}{c}{Head CT} & \\
  \cmidrule(lr){3-7}\cmidrule(lr){8-9}
  Method & Params & KiTS23 & H.\,vessel & Pancreas & PANORAMA & Colon & MSICH & INSTANCE22 & Mean \\
  \midrule
  \multicolumn{10}{l}{\textsc{trained from scratch}} \\
  UNETR-S & 33M & 17.7\std{4.4} & 23.6\std{3.1} & 8.9\std{4.2} & 5.4\std{2.9} & 2.3\std{1.4} & 30.2\std{7.8} & 44.2\std{16.5} & 18.9 \\
  UNETR-B & 99M & 21.4\std{6.8} & 22.8\std{4.2} & 8.4\std{2.9} & 6.7\std{4.2} & 2.3\std{1.8} & 30.7\std{8.5} & 45.1\std{19.7} & 19.6 \\
  Swin UNETR & 62M & 18.4\std{6.0} & 21.6\std{3.1} & 11.8\std{3.9} & 6.9\std{4.8} & 2.9\std{1.8} & 28.1\std{10.3} & 45.2\std{21.3} & 19.3 \\
  \midrule
  \multicolumn{10}{l}{\textsc{publicly released pretrained models}} \\
  Models Genesis & 19M & 16.7\std{6.3} & 23.2\std{9.3} & 11.4\std{5.7} & 5.4\std{3.3} & 3.4\std{6.3} & 36.5\std{8.1} & 50.6\std{26.5} & 21.0 \\
  Swin-SSL & 62M & 21.2\std{6.0} & 23.8\std{6.2} & 14.8\std{2.9} & 8.7\std{6.7} & 3.2\std{2.7} & 34.7\std{6.0} & 49.0\std{19.2} & 22.2 \\
  VoCo-B & 73M & 36.2\std{4.9} & 30.5\std{7.5} & 31.1\std{7.1} & 30.6\std{3.5} & 4.0\std{5.5} & 33.2\std{9.4} & 47.5\std{25.2} & 30.5 \\
  VoCo-L & 290M & 35.6\std{5.1} & 29.9\std{7.8} & 28.9\std{8.7} & 29.7\std{6.1} & 3.6\std{5.0} & 34.3\std{8.0} & 48.1\std{25.5} & 30.0 \\
  Hi-End-MAE & 361M & 42.4\std{8.1} & 29.8\std{7.0} & 33.5\std{5.4} & 30.4\std{9.3} & 6.3\std{9.5} & \underline{37.7}\std{9.5} & 47.1\std{26.6} & \underline{32.5} \\
  3DINO & 322M & 31.4\std{8.8} & 27.8\std{7.3} & 19.8\std{3.6} & 13.4\std{8.2} & 4.6\std{6.5} & 34.8\std{8.7} & 47.5\std{21.9} & 25.6 \\
  CT-FM & 87M & 25.2\std{9.6} & 24.9\std{7.9} & 13.9\std{5.2} & 9.6\std{4.2} & 2.8\std{4.2} & 34.6\std{8.6} & 48.1\std{27.1} & 22.7 \\
  Curia-2 & 100M & 38.7\std{3.5} & 30.1\std{6.8} & 31.3\std{6.8} & 27.8\std{8.6} & 4.9\std{9.6} & 35.3\std{8.1} & 46.2\std{24.9} & 30.6 \\
  \midrule
  \multicolumn{10}{l}{\textsc{gala-3d (ours)}} \\
  \textbf{GALA-S} & 33M & \underline{44.4}\std{8.6} & \underline{31.4}\std{8.5} & \underline{34.2}\std{5.5} & \textbf{34.0}\std{7.3} & \textbf{9.3}\std{10.7} & \textbf{37.8}\std{8.6} & \textbf{51.8}\std{25.7} & \textbf{34.7} \\
  \textbf{GALA-B} & 99M & \textbf{45.2}\std{8.4} & \textbf{32.2}\std{7.5} & \textbf{36.0}\std{4.8} & \underline{32.9}\std{7.7} & \underline{8.7}\std{10.0} & 36.5\std{9.2} & \underline{51.5}\std{23.9} & \textbf{34.7} \\
  \midrule
  GALA-S, head CT$^{\dagger}$ & 33M & -- & -- & -- & -- & -- & 40.1$^{\blacktriangle}$\std{8.4} & 52.5$^{\blacktriangle}$\std{22.9} & -- \\
  \bottomrule
\end{tabular*}

        \end{subtable}
        \par\medskip
        \begin{subtable}{\textwidth}
            \caption{1\% of the training pool}
            \label{tab:res-1pct}
            % Generated by downstream/NAB/plot/make_tex_tables.py; do not edit by hand.
\begin{tabular*}{\textwidth}{@{\extracolsep{\fill}}lr*{5}{c}c}
  \toprule
   & & \multicolumn{4}{c}{Abdominal CT} & \multicolumn{1}{c}{Head CT} & \\
  \cmidrule(lr){3-6}\cmidrule(lr){7-7}
  Method & Params & KiTS23 & H.\,vessel & Pancreas & PANORAMA & MSICH & Mean \\
  \midrule
  \multicolumn{8}{l}{\textsc{trained from scratch}} \\
  UNETR-S & 33M & 26.7\std{4.9} & 28.0\std{3.6} & 12.1\std{6.0} & 12.3\std{3.8} & 41.1\std{6.7} & 24.1 \\
  UNETR-B & 99M & 28.3\std{1.4} & 28.1\std{3.6} & 12.5\std{3.6} & 15.6\std{2.4} & 40.4\std{5.6} & 24.9 \\
  Swin UNETR & 62M & 28.1\std{2.7} & 27.2\std{3.5} & 14.6\std{3.0} & 15.7\std{3.6} & 40.3\std{6.6} & 25.2 \\
  \midrule
  \multicolumn{8}{l}{\textsc{publicly released pretrained models}} \\
  Models Genesis & 19M & 29.9\std{8.6} & 29.5\std{13.0} & 12.9\std{4.3} & 16.5\std{6.1} & 44.6\std{5.9} & 26.7 \\
  Swin-SSL & 62M & 28.9\std{3.5} & 31.0\std{5.5} & 16.4\std{4.2} & 20.3\std{3.5} & 43.5\std{5.0} & 28.0 \\
  VoCo-B & 73M & 42.2\std{3.7} & 39.1\std{6.3} & 36.0\std{3.6} & 42.8\std{3.8} & 44.2\std{5.6} & 40.9 \\
  VoCo-L & 290M & 43.4\std{4.1} & 39.2\std{6.2} & 34.4\std{3.1} & 44.8\std{4.3} & 43.5\std{5.0} & 41.1 \\
  Hi-End-MAE & 361M & 52.6\std{6.7} & 39.2\std{7.0} & \underline{39.1}\std{3.3} & \underline{45.6}\std{4.8} & 47.9\std{6.0} & 44.9 \\
  3DINO & 322M & 41.8\std{6.1} & 36.8\std{5.9} & 23.7\std{4.0} & 28.5\std{4.4} & 45.0\std{6.1} & 35.2 \\
  CT-FM & 87M & 37.9\std{8.3} & 34.7\std{7.5} & 17.9\std{5.3} & 23.4\std{5.0} & 46.5\std{5.2} & 32.1 \\
  Curia-2 & 100M & 46.7\std{6.9} & 38.5\std{5.9} & 37.0\std{5.9} & 42.5\std{2.4} & \underline{48.2}\std{5.5} & 42.6 \\
  \midrule
  \multicolumn{8}{l}{\textsc{gala-3d (ours)}} \\
  \textbf{GALA-S} & 33M & \underline{55.5}\std{6.3} & \textbf{43.3}\std{5.1} & 38.8\std{2.3} & 44.8\std{5.0} & 47.9\std{7.1} & \underline{46.1} \\
  \textbf{GALA-B} & 99M & \textbf{58.4}\std{7.4} & \underline{42.7}\std{5.8} & \textbf{40.5}\std{1.3} & \textbf{49.7}\std{2.3} & \textbf{48.7}\std{6.4} & \textbf{48.0} \\
  \midrule
  GALA-S, head CT$^{\dagger}$ & 33M & -- & -- & -- & -- & 52.9$^{\blacktriangle}$\std{6.7} & -- \\
  \bottomrule
\end{tabular*}

        \end{subtable}
    \end{table*}

    \begin{table*}[!t]
        \caption{\textbf{Held-out-test DSC (\%) of every method on the CT datasets with 10\% and 100\% of the labels.} Mean $\pm$ standard deviation over five folds. Mean: average over the datasets. Bold marks the best method in each column and underline the second best. $^{\dagger}$GALA-S pretrained on RSNA 2019 head CT with MSICH patients removed (\ref{app:headct}), shown for reference on the head-CT datasets and not ranked. $^{\blacktriangle}$Higher than every other method in the column.}
        \label{tab:res-high}
        \centering
        \footnotesize
        \setlength{\tabcolsep}{3pt}
        \captionsetup[subtable]{skip=2pt}
        \begin{subtable}{\textwidth}
            \caption{10\% of the training pool}
            \label{tab:res-10pct}
            % Generated by downstream/NAB/plot/make_tex_tables.py; do not edit by hand.
\begin{tabular*}{\textwidth}{@{\extracolsep{\fill}}lr*{7}{c}c}
  \toprule
   & & \multicolumn{5}{c}{Abdominal CT} & \multicolumn{2}{c}{Head CT} & \\
  \cmidrule(lr){3-7}\cmidrule(lr){8-9}
  Method & Params & KiTS23 & H.\,vessel & Pancreas & PANORAMA & Colon & MSICH & INSTANCE22 & Mean \\
  \midrule
  \multicolumn{10}{l}{\textsc{trained from scratch}} \\
  UNETR-S & 33M & 52.6\std{9.3} & 45.6\std{3.0} & 33.9\std{2.5} & 40.8\std{4.8} & 9.5\std{1.6} & 58.0\std{2.8} & 61.9\std{6.9} & 43.2 \\
  UNETR-B & 99M & 52.0\std{7.5} & 46.9\std{2.2} & 37.9\std{2.0} & 41.0\std{3.6} & 8.1\std{2.6} & 57.6\std{2.6} & 61.4\std{4.9} & 43.6 \\
  Swin UNETR & 62M & 52.0\std{8.3} & 44.4\std{1.5} & 38.6\std{1.0} & 46.2\std{5.7} & 12.4\std{2.9} & 59.9\std{2.3} & 63.7\std{6.2} & 45.3 \\
  \midrule
  \multicolumn{10}{l}{\textsc{publicly released pretrained models}} \\
  Models Genesis & 19M & 55.0\std{1.6} & 48.9\std{2.5} & 43.8\std{4.0} & 51.1\std{0.6} & 13.7\std{6.9} & 59.9\std{2.8} & 67.0\std{5.9} & 48.5 \\
  Swin-SSL & 62M & 54.7\std{7.7} & 46.6\std{3.4} & 38.6\std{3.4} & 47.7\std{5.1} & 11.8\std{2.3} & 60.8\std{2.1} & 65.5\std{6.2} & 46.5 \\
  VoCo-B & 73M & 70.9\std{3.4} & 55.2\std{2.7} & 54.0\std{2.2} & \underline{65.4}\std{1.6} & 27.7\std{6.6} & 61.2\std{3.3} & 66.0\std{4.7} & 57.2 \\
  VoCo-L & 290M & 72.1\std{3.1} & 55.7\std{1.1} & 53.2\std{3.0} & 64.0\std{0.8} & 28.0\std{6.8} & 62.3\std{4.1} & 66.9\std{5.1} & 57.5 \\
  Hi-End-MAE & 361M & 74.1\std{1.1} & 55.3\std{1.2} & 54.3\std{2.4} & 62.9\std{2.0} & 27.3\std{4.6} & 63.5\std{3.0} & 64.7\std{4.7} & 57.4 \\
  3DINO & 322M & 65.0\std{2.4} & 53.5\std{1.4} & 44.4\std{2.7} & 57.7\std{1.5} & 19.6\std{3.1} & 58.7\std{3.2} & 62.7\std{5.4} & 51.7 \\
  CT-FM & 87M & 62.1\std{5.3} & 53.3\std{0.9} & 46.3\std{3.1} & 54.3\std{4.2} & 21.5\std{9.0} & 62.1\std{3.7} & \textbf{68.7}\std{5.4} & 52.6 \\
  Curia-2 & 100M & 69.0\std{2.0} & 55.9\std{2.0} & \underline{55.2}\std{2.1} & 62.8\std{1.9} & 24.4\std{7.3} & \textbf{64.8}\std{1.4} & 65.9\std{3.8} & 56.9 \\
  \midrule
  \multicolumn{10}{l}{\textsc{gala-3d (ours)}} \\
  \textbf{GALA-S} & 33M & \textbf{75.8}\std{1.7} & \textbf{56.7}\std{1.9} & 54.4\std{1.9} & 65.1\std{2.2} & \underline{32.4}\std{5.7} & 62.4\std{2.4} & 67.3\std{4.6} & \underline{59.2} \\
  \textbf{GALA-B} & 99M & \underline{75.7}\std{2.2} & \underline{56.2}\std{0.9} & \textbf{56.8}\std{2.2} & \textbf{66.8}\std{1.5} & \textbf{35.4}\std{3.6} & \underline{64.0}\std{3.0} & \underline{67.7}\std{5.3} & \textbf{60.4} \\
  \midrule
  GALA-S, head CT$^{\dagger}$ & 33M & -- & -- & -- & -- & -- & 66.3$^{\blacktriangle}$\std{2.6} & 67.8\std{4.4} & -- \\
  \bottomrule
\end{tabular*}

        \end{subtable}
        \par\medskip
        \begin{subtable}{\textwidth}
            \caption{Full training pool}
            \label{tab:res-full}
            % Generated by downstream/NAB/plot/make_tex_tables.py; do not edit by hand.
\begin{tabular*}{\textwidth}{@{\extracolsep{\fill}}lr*{7}{c}c}
  \toprule
   & & \multicolumn{5}{c}{Abdominal CT} & \multicolumn{2}{c}{Head CT} & \\
  \cmidrule(lr){3-7}\cmidrule(lr){8-9}
  Method & Params & KiTS23 & H.\,vessel & Pancreas & PANORAMA & Colon & MSICH & INSTANCE22 & Mean \\
  \midrule
  \multicolumn{10}{l}{\textsc{trained from scratch}} \\
  UNETR-S & 33M & 70.4\std{9.8} & 58.3\std{4.1} & 58.9\std{1.8} & 70.1\std{1.6} & 32.7\std{6.6} & 71.8\std{2.3} & 70.6\std{3.3} & 61.8 \\
  UNETR-B & 99M & 79.0\std{3.3} & 59.4\std{1.5} & 58.7\std{1.7} & 64.3\std{6.7} & 29.0\std{6.9} & 71.1\std{2.4} & 71.2\std{4.1} & 61.8 \\
  Swin UNETR & 62M & 76.6\std{3.7} & 58.9\std{3.2} & 61.8\std{2.0} & 67.3\std{3.9} & 32.5\std{6.4} & \underline{73.3}\std{2.1} & 71.5\std{3.2} & 63.1 \\
  \midrule
  \multicolumn{10}{l}{\textsc{publicly released pretrained models}} \\
  Models Genesis & 19M & 74.4\std{2.7} & 59.3\std{1.0} & 54.7\std{2.9} & 65.3\std{2.9} & 33.3\std{5.7} & 70.7\std{2.2} & 73.5\std{3.7} & 61.6 \\
  Swin-SSL & 62M & 74.5\std{12.9} & 58.7\std{2.6} & 61.4\std{3.1} & 66.8\std{8.4} & 31.5\std{6.3} & 72.7\std{1.8} & 73.4\std{4.1} & 62.7 \\
  VoCo-B & 73M & 82.1\std{1.9} & 63.5\std{1.0} & 65.0\std{1.0} & \underline{73.1}\std{1.8} & 46.4\std{4.3} & 73.0\std{1.6} & 72.9\std{3.1} & 68.0 \\
  VoCo-L & 290M & 82.4\std{2.3} & 63.4\std{1.5} & \underline{66.0}\std{1.4} & 72.9\std{1.8} & 49.0\std{3.7} & \underline{73.3}\std{1.1} & 72.4\std{2.8} & 68.5 \\
  Hi-End-MAE & 361M & 82.9\std{1.9} & 62.3\std{2.2} & 63.6\std{2.2} & 71.8\std{2.5} & 42.4\std{1.6} & 72.9\std{2.2} & 71.5\std{3.2} & 66.8 \\
  3DINO & 322M & 77.0\std{3.9} & 61.7\std{2.5} & 58.3\std{2.4} & 70.3\std{2.0} & 35.8\std{4.1} & 70.9\std{1.6} & 70.0\std{3.1} & 63.4 \\
  CT-FM & 87M & 80.2\std{1.9} & 60.7\std{1.7} & 61.8\std{1.6} & 68.8\std{2.9} & 39.9\std{7.0} & 72.1\std{2.3} & 72.6\std{3.8} & 65.1 \\
  Curia-2 & 100M & 81.7\std{2.9} & \underline{63.9}\std{1.8} & 65.2\std{1.7} & 71.6\std{3.3} & 45.4\std{4.3} & \textbf{73.8}\std{1.9} & \underline{74.1}\std{2.8} & 68.0 \\
  \midrule
  \multicolumn{10}{l}{\textsc{gala-3d (ours)}} \\
  \textbf{GALA-S} & 33M & \textbf{84.5}\std{0.6} & 63.6\std{2.3} & \underline{66.0}\std{1.2} & \textbf{73.6}\std{1.8} & \underline{49.7}\std{3.3} & 73.2\std{2.3} & \textbf{74.4}\std{3.5} & \underline{69.3} \\
  \textbf{GALA-B} & 99M & \underline{83.8}\std{1.3} & \textbf{64.1}\std{1.9} & \textbf{66.4}\std{1.9} & \underline{73.1}\std{1.4} & \textbf{50.9}\std{3.7} & 73.1\std{2.5} & 74.0\std{3.8} & \textbf{69.4} \\
  \midrule
  GALA-S, head CT$^{\dagger}$ & 33M & -- & -- & -- & -- & -- & 72.9\std{1.9} & 73.3\std{4.4} & -- \\
  \bottomrule
\end{tabular*}

        \end{subtable}
    \end{table*}

    \begin{figure*}[!tp]
        \centering
        \includegraphics[width=\textwidth]{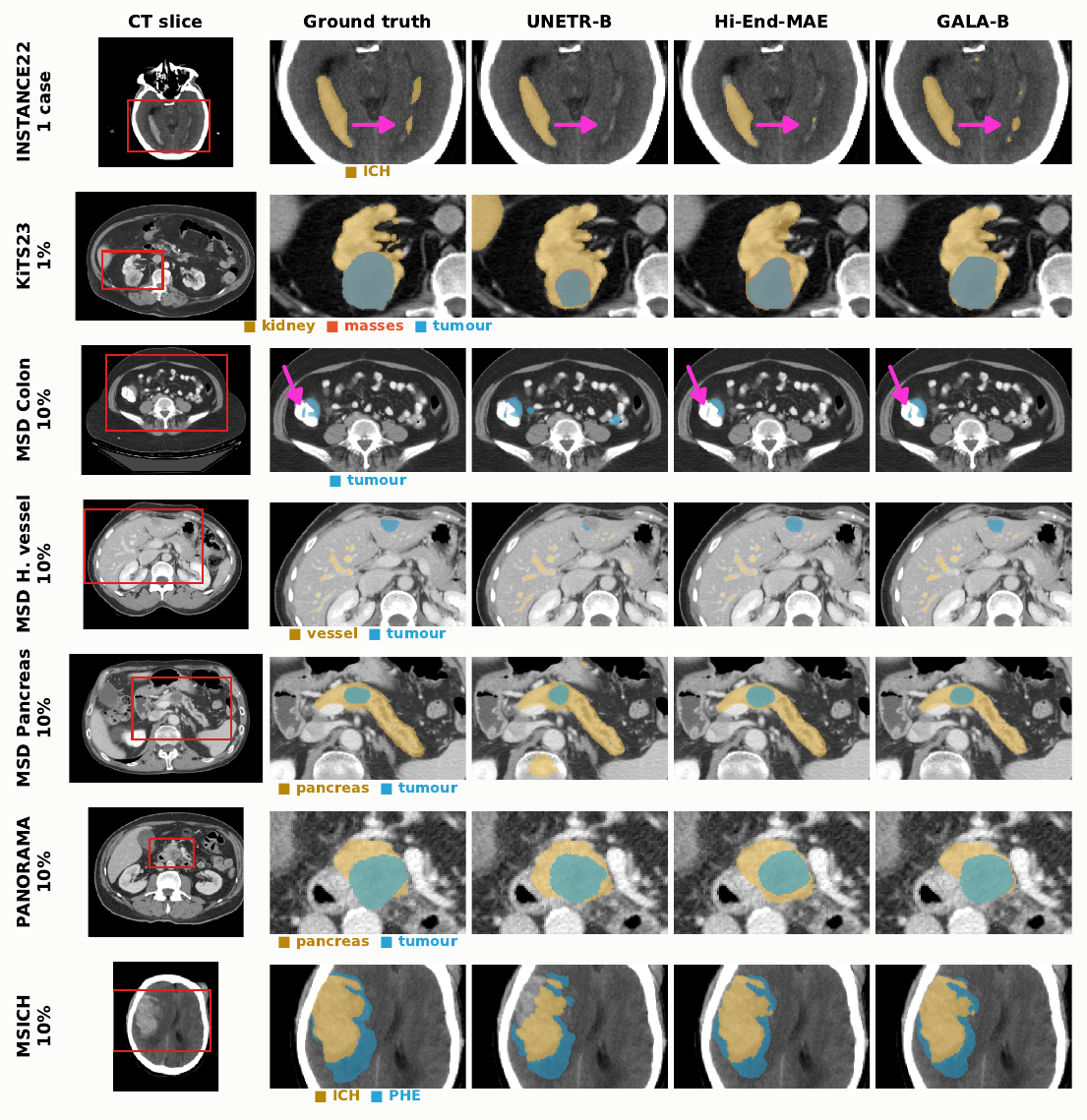}
        \caption{\textbf{Qualitative comparison at low label budgets, ordered by budget.} INSTANCE22 with one labelled case, KiTS23 with 1\%, and the five other CT datasets with 10\% of the training labels. Left: full slice, with the red box marking the region shown in the other columns. Arrows mark a small separate haemorrhage on INSTANCE22 and a small separate part of the tumour on MSD Colon, both detected by GALA-B. Cases were selected to illustrate the DSC gain of GALA-B over Hi-End-MAE, the external model with the highest overall mean DSC. Aggregate results are given in \ref{app:results}.}
        \label{fig:qualitative}
    \end{figure*}

    \subsection{Downstream segmentation on CT}
    \label{sec:downstream-tasks}

    Figures~\ref{fig:budget-bars} and~\ref{fig:label-efficiency} and Tables~\ref{tab:res-low} and~\ref{tab:res-high} summarize the comparison on the seven CT datasets, and the paired tests and further results are given in \ref{app:results}.

    \paragraph{Comparison with external pretrained models} At every label budget, the two GALA-3D encoders obtain the two highest mean DSC values among the thirteen methods (Figure~\ref{fig:budget-bars}). The same ordering holds for ranks. When the methods are ranked by mean DSC on each dataset, from 1 (best) to 13, a GALA-3D encoder obtains the best mean rank at every budget (1.2 to 2.1), ahead of the best external model (2.8 to 4.4). The pairwise comparisons support this advantage. GALA-B exceeds every external model in mean DSC at every budget, significantly in 30 of the 32 paired comparisons after Holm correction (Table~\ref{tab:stats-vitb8}). The two exceptions, VoCo-L and Curia-2, occur with the full training pool, where the strongest initializations converge. GALA-S, with a quarter of the encoder parameters, likewise exceeds every external model at every budget, significantly in 29 of 32 comparisons (Table~\ref{tab:stats-vits8}).

    \paragraph{Margins per setting} Tables~\ref{tab:res-low} and~\ref{tab:res-high} report every dataset and label budget. Against Hi-End-MAE, the external model with the highest overall mean DSC, GALA-B is best or tied in 25 of the 26 CT settings, counting differences within 0.1 points as ties. Against the strongest external model chosen separately in each setting, a stricter reference, it is best or tied in 16 of the 19 settings between one labelled case and 10\% of the labels, with a mean margin of 2.0 DSC points. The gains also persist under the two-level bootstrap (Section~\ref{sec:evaluation}), which places the 95\% interval of the mean DSC difference above zero for every comparison of GALA-B with an external model at every budget, and the interval only approaches zero for VoCo-L and Curia-2 with the full training pool, the two comparisons that are not significant after Holm correction. Finally, prior exposure to the evaluation data does not account for the margins. On KiTS23, whose cases are included in the pretraining corpora of VoCo, Hi-End-MAE, 3DINO, and CT-FM but not in that of GALA-3D (Table~\ref{tab:overlap}), GALA-B outperforms every external model at all four label budgets.

    \paragraph{Boundary agreement} Boundary agreement, measured by NSD, leads to the same conclusions as DSC. GALA-B exceeds every external model at every budget, significantly in 30 of the 32 comparisons, with the same two exceptions (Table~\ref{tab:nsd-vitb8}). This advantage does not depend on the chosen tolerance. With tolerances of 1 and 5\,mm, GALA-B remains ahead of all eight external models at every budget, except VoCo-L at 1\,mm with the full training pool (Table~\ref{tab:nsd-tolerance}).

    \paragraph{Label efficiency} The benefit of pretraining is largest when annotations are scarcest (Figure~\ref{fig:label-efficiency}). With a single labelled case, GALA-B reaches 45.2\% DSC on KiTS23, more than twice the 21.4\% of UNETR-B, the same network trained from scratch. With 10\% of the labels, it already exceeds UNETR-S trained on the full training pool on KiTS23 and MSD Colon. The benefit narrows but does not disappear with all labels. GALA-B then exceeds both UNETR-B and UNETR-S in each of the 35 (dataset, fold) pairs, by 7.5 points on average.

    \paragraph{Qualitative results} Figure~\ref{fig:qualitative} shows test cases at low label budgets, selected to illustrate the DSC gain of GALA-B over Hi-End-MAE, the external model with the highest overall mean DSC. On INSTANCE22 with a single labelled case, GALA-B recovers a small separate haemorrhage, marked by the arrow, which UNETR-B misses and Hi-End-MAE only partly detects. On KiTS23, UNETR-B segments only part of the kidney tumour, Hi-End-MAE recovers more of it, and GALA-B follows its boundary most closely. On MSD Colon, UNETR-B finds the tumour but also predicts false-positive tumour at two other sites in the bowel, whereas Hi-End-MAE misses a small separate part of the tumour, marked by the arrow, that GALA-B detects. On MSD hepatic vessel, UNETR-B detects only a small part of the liver tumour, whereas both pretrained models find it and GALA-B outlines it more tightly. On MSD Pancreas, UNETR-B under-segments the tumour and Hi-End-MAE over-extends it, while GALA-B matches its extent most closely. On PANORAMA, UNETR-B and Hi-End-MAE over-segment the pancreas around the tumour, whereas GALA-B matches the extent of both structures most closely. On MSICH, UNETR-B misses a large part of the haemorrhage, which GALA-B extends up to the skull. Across these cases, training from scratch fails mainly on lesions, which it misses, under-segments, or confuses with neighbouring tissue. Pretraining avoids most of these failures, and the two pretrained models differ mainly in how closely they delineate lesions, with GALA-B matching lesion boundaries more closely and detecting small separate lesions that the other models miss.

    \paragraph{In-domain and out-of-domain CT} The advantage is most pronounced on abdominal CT, the domain of the pretraining cohort. On the five abdominal datasets, GALA-B is best or tied in all 19 settings across the four label budgets, with a mean margin of 2.2 points over the strongest external model and up to 7.4 points on MSD Colon with 10\% of the labels (Table~\ref{tab:res-10pct}). This margin comes mainly from lesions. Against the strongest external model for each class, GALA-B is ahead or tied in 16 of the 17 abdominal lesion class--budget settings between one labelled case and 10\% of the labels, by 3.9 points on average, whereas on the organ and vessel classes the leading models are on par (Table~\ref{tab:class-margins}).

    Head CT is a harder test, since it is absent from the GALA-3D pretraining cohort but present in that of several external models (Table~\ref{tab:overlap}). Even so, GALA-B is best or tied in three of the seven head CT settings and within 1.2 points of the strongest external model in the other four, and both GALA-3D encoders improve on their size-matched scratch counterparts by 1.4 to 8.3 points. No single external model leads on head CT. Moreover, when GALA-S is pretrained on head CT, it exceeds the strongest external model in four of the seven head CT settings, including every limited budget on MSICH, and remains within 0.9 points in the other three (\ref{app:headct}).

    \subsection{Ablation experiments}
    \label{sec:ablations}

    Single-factor ablations change one setting of GALA-S at a time, with the same five folds and fine-tuning protocol as the main comparison. They are evaluated on KiTS23, the largest of the evaluated datasets, which combines an organ with two lesion classes, at two label budgets, one labelled case and 10\% of the training pool, that bracket the low-label range. The component study (Table~\ref{tab:ablation-components}) adds the components of GALA-3D one at a time, starting from LeJEPA with independently sampled views, and removes each of them in turn from GALA-S. The sensitivity study then varies the number of local views, the projection dimension, the SIGReg weights $\lambda_g$ and $\lambda_d$, the mask ratio, and the patch edge. The comparison with MAE3D, which concerns the objective as a whole, is evaluated on all seven CT datasets (Section~\ref{sec:objective-control}).

    \paragraph{Components} Every component contributes to the default GALA-S (Table~\ref{tab:ablation-components}). Building up from LeJEPA, which reaches 48.2 DSC averaged over the two budgets, the anatomy-aware shared parent raises DSC by 6.7 points to 54.9. Adding dense latent prediction without its regularizer makes the dense representation collapse (19.8), and adding the patch-level SIGReg recovers it and raises DSC to 60.1. Conversely, removing a single component from GALA-S costs 4.2 points without the shared parent region, 4.5 without the global branch, and 5.8 without the global SIGReg. The two branches are therefore complementary, and each needs its own regularizer.

    \paragraph{Dense collapse} Without the patch-level SIGReg, or with its weight reduced to $\lambda_d=0.001$, the scale of the patch tokens becomes unstable during training, and the final patch tokens collapse towards a positional code. For the model trained without the patch-level SIGReg, patch position explains 28\% of the variance of the final patch tokens on 256 held-out abdominal CT crops (the share carried by the per-position mean token), against 3\% for the default GALA-S, and their effective rank is about a fifth of that of the default GALA-S (16.4 against 75.4). The predictor, which is conditioned on the same coordinates, can then match the targets without using image content.

    \paragraph{Sensitivity} Outside the collapse regime, GALA-3D is robust to most of its hyperparameters (Table~\ref{tab:sensitivity}). No explored alternative improves the average over the two budgets by more than 0.6 points, well below the fold-to-fold standard deviation. Projection dimensions from 64 to 256, global SIGReg weights from 0.01 to 0.1, a dense weight of 0.1, mask ratios from 0.75 to 0.95, and removing the local views change DSC by at most 1.9 points. Reducing the dense weight to $\lambda_d=0.01$ lowers DSC more, by 1.6 and 3.7 points, between the default and the collapse at $\lambda_d=0.001$. More extreme settings mainly lower the 10\% result, by 2.1 to 4.4 points for a projection dimension of 512, a mask ratio of 0.5, and $\lambda_g=0.001$. The patch size has the largest effect, as patches of 12 and 16 voxels reduce DSC at 10\% by 3.1 and 7.8 points, consistent with the importance of fine spatial detail and of the compression ratio of the patch embedding for segmentation.

    \begin{table*}[!t]
        \caption{\textbf{Component analysis of GALA-S on KiTS23.} Held-out-test DSC (\%), mean $\pm$ standard deviation over five folds. Avg: mean of the two budgets. $\Delta$: change in Avg relative to the row above in (a) and to GALA-S in (b). $^{\dagger}$Collapse of the dense representation without its regularizer.}
        \label{tab:ablation-components}
        \vspace{5pt}
        \centering
        \footnotesize
        \setlength{\tabcolsep}{4pt}
        % Generated by downstream/NAB/plot/make_tex_tables.py; do not edit by hand.
\begin{tabular}{lccccccccc}
  \toprule
  Configuration & \makecell{Global\\branch} & \makecell{Global\\SIGReg} & \makecell{Shared\\parent} & \makecell{Dense\\branch} & \makecell{Patch\\SIGReg} & 1 case & 10\% & Avg & $\Delta$ \\
  \midrule
  \multicolumn{10}{l}{\textit{(a) Adding components, starting from LeJEPA}} \\
  LeJEPA (independent views) & \checkmark & \checkmark &  &  &  & 31.9\std{3.0} & 64.5\std{3.1} & 48.2 & -- \\
  + anatomy-aware shared parent & \checkmark & \checkmark & \checkmark &  &  & 40.0\std{3.6} & 69.9\std{2.7} & 54.9 & $+6.7$ \\
  + dense latent prediction & \checkmark & \checkmark & \checkmark & \checkmark &  & 14.2\std{3.1} & 25.4\std{10.8} & 19.8$^{\dagger}$ & $-35.1$ \\
  + patch-level SIGReg (GALA-S) & \checkmark & \checkmark & \checkmark & \checkmark & \checkmark & \textbf{44.4}\std{8.6} & \textbf{75.8}\std{1.7} & \textbf{60.1} & $+40.3$ \\
  \midrule
  \multicolumn{10}{l}{\textit{(b) Removing one component from GALA-S}} \\
  $-$ shared parent (independent views) & \checkmark & \checkmark &  & \checkmark & \checkmark & 41.0\std{3.5} & 70.7\std{3.3} & 55.9 & $-4.2$ \\
  $-$ global SIGReg & \checkmark &  & \checkmark & \checkmark & \checkmark & 39.6\std{3.9} & 68.9\std{3.6} & 54.3 & $-5.8$ \\
  $-$ global branch (dense only) &  &  & \checkmark & \checkmark & \checkmark & 41.3\std{5.6} & 69.9\std{1.7} & 55.6 & $-4.5$ \\
  \bottomrule
\end{tabular}

    \end{table*}

    \begin{table}[!t]
        \footnotesize
        \caption{\textbf{Hyperparameter sensitivity of GALA-S on KiTS23.} Entries are held-out-test mean DSC (\%) over five folds, $\pm$ the standard deviation across folds. The default value of each factor is given in parentheses. Average: mean of the two budgets.}
        \label{tab:sensitivity}
        \vspace{5pt}
        \centering
        \setlength{\tabcolsep}{1.7pt}
        \begin{adjustbox}{max width=\columnwidth}
            % Generated by downstream/NAB/plot/make_tex_tables.py; do not edit by hand.
\begin{tabular}{l>{\centering\arraybackslash}p{2.5cm}ccc}
  \toprule
  Factor & Setting & 1 case & 10\% & Average \\
  \midrule
  Default & GALA-S, default settings & 44.4\std{8.6} & 75.8\std{1.7} & 60.1 \\
  \midrule
  Local views $V_l$ (2) & 0 & 43.8\std{6.0} & 74.6\std{2.3} & 59.2 \\
  \addlinespace
  Projection dimension $k$ (128) & 64 & 44.6\std{6.3} & 75.8\std{1.6} & 60.2 \\
   & 256 & 45.5\std{6.5} & 76.0\std{2.2} & 60.7 \\
   & 512 & 44.5\std{7.1} & 73.7\std{2.4} & 59.1 \\
  \addlinespace
  Global $\lambda_g$ (0.05) & 0.001 & 43.8\std{7.0} & 71.4\std{3.4} & 57.6 \\
   & 0.01 & 45.3\std{5.5} & 75.1\std{2.7} & 60.2 \\
   & 0.1 & 45.5\std{6.1} & 75.0\std{2.6} & 60.3 \\
  \addlinespace
  Dense $\lambda_d$ (0.05) & 0.001 & 13.0\std{2.2} & 23.9\std{6.4} & 18.4 \\
   & 0.01 & 42.8\std{4.3} & 72.1\std{2.8} & 57.4 \\
   & 0.1 & 44.1\std{5.9} & 75.2\std{2.5} & 59.7 \\
  \addlinespace
  Mask ratio $\rho$ (0.95) & 0.50 & 44.1\std{5.2} & 73.3\std{2.4} & 58.7 \\
   & 0.75 & 43.0\std{6.4} & 75.3\std{2.5} & 59.1 \\
   & 0.85 & 44.1\std{6.2} & 75.6\std{2.7} & 59.8 \\
   & 0.90 & 46.3\std{6.0} & 75.0\std{2.6} & 60.6 \\
  \addlinespace
  Patch edge (8) & 12 & 44.2\std{4.3} & 72.7\std{4.0} & 58.4 \\
   & 16 & 39.1\std{4.1} & 68.0\std{2.9} & 53.5 \\
  \bottomrule
\end{tabular}

        \end{adjustbox}
    \end{table}

    \begin{table}[!tb]
        \footnotesize
        \caption{\textbf{Pretraining-objective ablation.} GALA-3D against MAE3D with the same encoder, pretraining cohort, and update budget. Mean held-out-test DSC (\%) over the paired (dataset, fold) cells of the seven CT datasets (five at 1\%), and their difference $\Delta$ (points). Bold marks the higher of each pair. Two-sided Wilcoxon signed-rank test, Holm-corrected within each encoder's four comparisons: $^{*}p<0.05$, $^{**}p<0.01$.}
        \label{tab:matched}
        \vspace{5pt}
        \centering
        \setlength{\tabcolsep}{2pt}
        \begin{adjustbox}{max width=\columnwidth}
            % Generated by downstream/NAB/plot/make_paper_stats.py; do not edit by hand.
\begin{tabular}{ccccccc}
  \toprule
  & \multicolumn{3}{c}{ViT-S/8} & \multicolumn{3}{c}{ViT-B/8} \\
  \cmidrule(lr){2-4} \cmidrule(lr){5-7}
  Labels & MAE-S & GALA-S & $\Delta$ & MAE-B & GALA-B & $\Delta$ \\
  \midrule
  1 case & 29.5 & \textbf{34.7} & +5.2$^{**}$ & 30.0 & \textbf{34.7} & +4.7$^{**}$ \\
  1\% & 40.3 & \textbf{46.1} & +5.8$^{**}$ & 42.8 & \textbf{48.0} & +5.2$^{**}$ \\
  10\% & 56.9 & \textbf{59.2} & +2.3$^{**}$ & 57.9 & \textbf{60.4} & +2.5$^{**}$ \\
  100\% & 68.9 & \textbf{69.3} & +0.4 & 69.3 & \textbf{69.4} & +0.1 \\
  \bottomrule
\end{tabular}

        \end{adjustbox}
    \end{table}

    \subsection{Pretraining objective}
    \label{sec:objective-control}

    The external models differ from GALA-3D in pretraining data as well as in objective. The same-data MAE3D control removes the first difference and isolates the pretraining framework, on the seven CT datasets, at every label budget and for both encoder sizes (Table~\ref{tab:matched}). GALA-3D exceeds MAE3D of the same size at every label budget, significantly from one labelled case to 10\% of the labels for both encoders. The margin is largest at 1\% (5.8 and 5.2 points for GALA-S and GALA-B) and with one labelled case (5.2 and 4.7 points), and it decreases to 2.3 to 2.5 points at 10\%. With the full training pool, the two objectives are on par ($+0.4$ and $+0.1$ points), as the strongest initializations converge.

    Encoder size matters less than the pretraining objective. Most of the benefit of GALA-3D is already obtained with the 21M-parameter GALA-S. Moving to GALA-B adds 1.9 points at 1\% and 1.2 points at 10\% of the labels, both significant, and leaves mean DSC nearly unchanged with one labelled case and at 100\% (Table~\ref{tab:stats-vitb8}). Scaling MAE3D helps little as well, as MAE-B exceeds MAE-S by only 0.4 to 2.5 points. The objective, in contrast, makes a large difference. GALA-S exceeds MAE-B, which has four times as many encoder parameters, by 4.7, 3.3, and 1.3 points with one labelled case, 1\%, and 10\% of the labels, and matches it with the full training pool.

    \subsection{Adaptation to brain MRI}
    \label{sec:mri}

    ISLES22 is the most distant transfer setting of this evaluation, since neither brain anatomy nor MRI contrast is represented in the CT-only MERLIN cohort. The two MRI channels also require an adapted input stem. For every pretrained encoder, the single-channel stem kernel is replicated across the two channels and divided by two, so that identical channels reproduce the response of the pretrained stem, and all other pretrained weights are loaded unchanged.

    Even under this double shift in anatomy and modality, pretraining on CT transfers to MRI. Both GALA-3D encoders improve on their size-matched scratch counterparts at every label budget, by up to 9.6 points with one labelled case for GALA-B (Table~\ref{tab:isles22}). With one labelled case, GALA-B is also the second-best of the fifteen methods, behind only Models Genesis. Moreover, both GALA-3D encoders exceed Hi-End-MAE, the strongest external model on CT, at every budget, and they exceed 3DINO at every budget even though 3DINO was pretrained with MRI. GALA-B also leads Curia-2, the other model pretrained with MRI, with one labelled case, and closes the gap at 1\% (55.8 against 55.9).

    With additional labels, the best model on brain MRI is CT-FM at 1\%, Models Genesis and Curia-2 at 10\%, and Curia-2 with the full training pool, and GALA-B trails the best model by 1.1 to 4.0 points over these budgets. This gap is not specific to the GALA-3D objective, since GALA-3D and MAE3D of the same size differ by at most 1.4 points at every budget, except with one labelled case, where GALA-B leads MAE-B by 3.4 points. We instead hypothesize that it reflects the pretraining domain. The head-CT experiment supports this hypothesis. MERLIN contains no head images either, and pretraining GALA-S on head CT instead, with only its intensity windows adapted, makes it the best method on MSICH at every limited label budget (\ref{app:headct}). Possible causes specific to MRI are discussed in Section~\ref{sec:discussion}.

    \begin{table}[!tp]
        \caption{\textbf{Segmentation of ISLES22 brain MRI.} Entries are held-out-test mean DSC (\%) over five folds, $\pm$ the standard deviation across folds. Bold marks the best method at each budget and underline the second best. Params: parameters of the fine-tuned network, encoder and decoder. MRI: whether the pretraining data included MRI ($\checkmark$) or not ($\times$), ---: no pretraining.}
        \label{tab:isles22}
        \vspace{5pt}
        \centering
        \footnotesize
        \setlength{\tabcolsep}{2pt}
        \begin{adjustbox}{max width=\columnwidth}
            % Generated by downstream/NAB/plot/make_tex_tables.py; do not edit by hand.
\begin{tabular}{lrccccc}
  \toprule
  Method & Params & MRI & 1 case & 1\% & 10\% & 100\% \\
  \midrule
  UNETR-S & 33M & --- & 41.9\std{6.8} & 49.5\std{4.3} & 63.8\std{2.5} & 73.3\std{2.5} \\
  UNETR-B & 99M & --- & 39.6\std{6.5} & 47.2\std{2.8} & 63.9\std{3.6} & 73.9\std{2.8} \\
  Swin UNETR & 62M & --- & 39.8\std{17.6} & 54.1\std{4.3} & 66.9\std{3.0} & 75.9\std{1.2} \\
  \addlinespace
  MAE-S & 33M & $\times$ & 48.8\std{6.0} & 55.8\std{4.1} & 66.8\std{3.4} & 74.6\std{4.1} \\
  MAE-B & 99M & $\times$ & 45.8\std{9.3} & 54.5\std{5.9} & 65.9\std{4.0} & 75.3\std{2.4} \\
  \addlinespace
  Models Genesis & 19M & $\times$ & \textbf{52.1}\std{4.9} & 54.5\std{6.9} & \textbf{69.6}\std{2.6} & 75.0\std{2.8} \\
  Swin-SSL & 62M & $\times$ & 43.3\std{13.1} & 51.2\std{3.3} & 67.5\std{3.5} & 76.2\std{2.3} \\
  VoCo-B & 73M & $\times$ & 40.4\std{6.2} & 46.6\std{2.8} & 65.2\std{2.8} & 75.8\std{3.1} \\
  VoCo-L & 290M & $\times$ & 43.3\std{7.4} & 49.7\std{5.5} & 66.8\std{2.5} & \underline{76.6}\std{2.7} \\
  Hi-End-MAE & 361M & $\times$ & 43.1\std{9.4} & 50.7\std{3.9} & 63.8\std{4.8} & 73.1\std{3.2} \\
  3DINO & 322M & \checkmark & 38.2\std{9.4} & 46.8\std{3.0} & 59.6\std{4.4} & 71.9\std{3.2} \\
  CT-FM & 87M & $\times$ & 48.4\std{5.5} & \textbf{58.1}\std{5.4} & \underline{69.3}\std{2.9} & 75.3\std{3.2} \\
  Curia-2 & 100M & \checkmark & 45.9\std{8.9} & \underline{55.9}\std{6.1} & \textbf{69.6}\std{4.0} & \textbf{77.4}\std{2.4} \\
  \addlinespace
  GALA-S & 33M & $\times$ & 49.1\std{4.6} & 54.5\std{4.8} & 65.4\std{4.3} & 75.2\std{3.5} \\
  GALA-B & 99M & $\times$ & \underline{49.2}\std{5.2} & 55.8\std{4.2} & 65.6\std{4.8} & 76.3\std{3.0} \\
  \bottomrule
\end{tabular}

        \end{adjustbox}
    \end{table}

    \subsection{Computational cost}
    \label{sec:compute}

    We count the floating-point operations (FLOPs) of one pretraining step, forward and backward, for one image with its two global and two local views (\ref{app:pretraining}). Training only with the global branch costs 935 GFLOPs for ViT-S/8 and 3,012 GFLOPs for ViT-B/8. The dense branch adds 47 and 151 GFLOPs, an increase of 5\% for both encoders, because it encodes only the 86 visible tokens of each masked global view, out of 1,728, and queries 82 masked positions. A complete pretraining run therefore runs on a single NVIDIA A100 40\,GB GPU, in 30 hours for GALA-S and 60 hours for GALA-B (Section~\ref{sec:implementation-details}).

    \section{Discussion}
    \label{sec:discussion}
    These results show that teacher-free joint-embedding pretraining on a modest, single-source cohort can outperform models pretrained on far larger corpora. Pretrained on 25,414 abdominal CT images, GALA-3D exceeds eight publicly released pretrained models across seven CT segmentation datasets, most clearly when labels are scarce, and it also outperforms a masked autoencoder pretrained on the same images. The following sections examine the source of this advantage, its computational cost, and how far it transfers beyond abdominal CT.
    \paragraph{Source of the improvement} The component ablations show that both the global and dense branches are needed (Section~\ref{sec:ablations}), and the same-data MAE3D control attributes the advantage over masked reconstruction to the pretraining framework rather than to the data (Section~\ref{sec:objective-control}). We hypothesize that the two branches learn complementary structure. The global branch makes the representation of a region invariant to field of view, intensity window, and orientation, while the dense branch makes each patch embedding predictable from its anatomical context, which may give the segmentation decoder spatially organized features to build on. Consistent with this view, removing either branch lowers DSC by a similar margin (Table~\ref{tab:ablation-components}). Because both targets lie in latent space rather than in voxel intensities, the encoder also does not have to reconstruct acquisition noise or fine texture, which may explain part of the margin over MAE3D. These properties should matter most when the decoder is learned from one or a few labelled cases, in line with the larger gains at low label budgets, whereas with the full training set fine-tuning recovers much of what pretraining provides and the strongest initializations converge.

    \paragraph{Computational efficiency} Training GALA-3D costs only about 5\% more than training its global objective alone (Section~\ref{sec:compute}), yet adding the dense branch with its patch-level SIGReg raises DSC on KiTS23 from 54.9 to 60.1 averaged over the two label budgets (Table~\ref{tab:ablation-components}). The dense branch is therefore a favourable trade-off. Spatially resolved supervision does not require a heavier recipe: masking the context and subsampling the queries keep its cost marginal, and the whole model still trains on a single GPU.
    
    The data requirements are also modest. The pretraining cohort of 25,414 images from a single abdominal CT collection is about six times smaller than those of CT-FM and VoCo, yet the 21M-parameter GALA-S already exceeds every external model in mean DSC at every label budget. Together with the robustness of the recipe to most hyperparameters outside the collapse regime (Table~\ref{tab:sensitivity}), this suggests that GALA-3D can be pretrained on new CT collections without extensive tuning.

    \paragraph{Scope of transfer} Transfer performance decreases gradually with distance from the pretraining domain. On abdominal CT, GALA-3D is best or tied in every setting. On head CT, an anatomy absent from its pretraining data but present in that of several external models, it remains competitive with the strongest methods. When pretrained on head CT with only its intensity windows adapted, GALA-S becomes the best method on MSICH at every limited label budget (\ref{app:headct}). On brain MRI, GALA-3D improves on its size-matched scratch counterparts but trails the best external models once more labels are available, a gap shared by MAE3D (Section~\ref{sec:mri}). The gap does not follow from CT-only pretraining alone, since Models Genesis and CT-FM, the strongest models at low budgets, were also pretrained only on CT. Both are convolutional networks, whose stronger local inductive bias may make them less dependent on the pretraining domain than a ViT encoder, one of the architectural differences noted among the limitations below. 
    
    Furthermore, two additional factors are specific to the pretraining data and pipeline of GALA-3D. First, views are generated under randomized clinical HU windows (Section~\ref{sec:global-local-representation-learning}), an invariance that has no counterpart for the arbitrary intensity scale of diffusion MRI. Second, MERLIN contains no brain anatomy, whereas Curia-2, tied for the best result with 10\% of the labels and the best with the full training pool, was pretrained on clinical MRI as well as CT \citep{curia2}. As the objective itself makes no assumption about modality, pretraining on a mixed CT and MRI cohort with modality-appropriate intensity augmentation is a natural future extension.

    \paragraph{Limitations} Three limitations qualify these findings. First, all downstream evaluations concern segmentation with a UNETR decoder and full fine-tuning, so the behaviour of GALA-3D in classification, detection, and frozen-encoder transfer is not established. Second, each GALA-3D model is pretrained on a single CT cohort of about 25,000 images, abdominal or head, with encoders up to ViT-B/8, so its behaviour with larger, multi-region cohorts and larger encoders remains untested. Third, the external models are evaluated through their released weights and native backbones, so their comparison with GALA-3D reflects differences in data, architecture, and compute in addition to the objective. The same-data MAE3D control isolates the pretraining framework, but only with respect to the masked reconstruction family.

    \paragraph{Future work} A natural extension is to pretrain GALA-3D on larger and more diverse corpora that span multiple anatomical regions and modalities, with larger encoders, which would also test whether the transfer gap on brain MRI closes when MRI is represented during pretraining, as the gap on MSICH does when head CT is (\ref{app:headct}). Evaluating the learned representations beyond segmentation, including frozen-encoder transfer, would further characterize their generality. Finally, language-free self-supervised encoders trained at scale have been shown to match language-supervised ones as vision backbones for multimodal language models \citep{webssl}, and SIGReg-based image--text pretraining has yielded strong frozen backbones for such models \citep{levljepa}. Because MERLIN pairs every image with its radiology report, coupling GALA-3D with a language model for report generation or visual question answering is a promising direction.

    \section{Conclusion}
    \label{sec:conclusion}

    This study presented GALA-3D, a teacher-free self-supervised pretraining framework for 3D images. GALA-3D draws global and local views from a shared foreground-anchored parent region and combines global view invariance with dense masked latent prediction, both regularized by SIGReg, without negative pairs, a momentum teacher, or a voxel-space decoder. Teacher-free dense latent prediction was shown to collapse in the absence of a patch-level regularizer. With the proposed patch-level SIGReg, training remained stable and the global and dense objectives proved complementary.

    Pretrained on 25,414 abdominal CT images that share no data with any evaluation dataset, GALA-3D obtains the highest mean Dice among thirteen methods at every label budget on seven CT segmentation datasets, including eight publicly released pretrained medical models,  several of which had been exposed to some of the evaluation datasets during pretraining. The improvements are largest when labels are scarce, where GALA-B exceeds the strongest external model by 2.2 to 3.1 points from one labelled case to 10\% of the labels. A masked autoencoder pretrained on the same images with the same encoders trails GALA-3D by 2.3 to 5.8 points at these budgets, which attributes the gain over masked reconstruction to the pretraining framework rather than to the data.

    The framework also extends beyond its pretraining domain. Pretrained on head CT, with only its intensity windows adapted to the new anatomy, GALA-S becomes the best method on MSICH at every limited label budget, and on brain MRI, a modality absent from its pretraining data, CT-pretrained GALA-3D still improves on its size-matched scratch counterparts at every budget. These improvements come at a modest computational cost: relative to the global-only objective, the dense branch increases the cost of a training step by only 5\%, and a complete pretraining run fits on a single GPU.

    These findings support GALA-3D as a practical and label-efficient initialization for 3D image segmentation under full fine-tuning. Pretraining it on larger cohorts that span several anatomical regions and modalities is a natural next step.

    \section*{CRediT authorship contribution statement}
    \textbf{Aloys Portafaix:} Conceptualization, Methodology, Software, Investigation, Formal analysis, Data curation, Visualization, Writing -- original draft, Writing -- review \& editing. \textbf{Laurent L\'etourneau-Guillon:} Data curation, Resources, Supervision, Writing -- review \& editing. \textbf{Samuel Kadoury:} Conceptualization, Resources, Supervision, Writing -- review \& editing.

    \section*{Declaration of generative AI and AI-assisted technologies in the manuscript preparation process}
    
    In preparing this manuscript, the authors used Claude (Anthropic) to improve the clarity and quality of the writing, assist with data visualization code, and support the development of selected portions of the code. All AI-assisted text and code were reviewed and revised by the authors as needed. The authors take full responsibility for the accuracy, integrity, and content of the published article and its associated code.

    \section*{Declaration of competing interest} 
    The authors declare that they have no known competing financial interests or personal relationships that could have appeared to influence the work reported in this paper.

    \section*{Funding}
    Aloys Portafaix is supported by an NSERC Canada Graduate Scholarship--Doctoral (CGS D). 

    This project was partly supported by a Quebec Bio-Imaging Network 2023 pilot project grant.

    Laurent L\'etourneau-Guillon is supported by a Fonds de recherche du Qu\'ebec -- Sant\'e (FRQS)/Fondation de l'Association des radiologistes du Qu\'ebec (FARQ) Junior 2 salary award (doi.org/10.69777/380236). 

    The funding sources had no involvement in the study design, the collection, analysis and interpretation of data, the writing of the report, or the decision to submit the article for publication.

    \section*{Ethics statement}
    This study uses only publicly available datasets, accessed under their respective licences, and no new patient data were collected for this work. Sex and gender were not analysed, as they are not consistently reported across the public datasets.

    \section*{Data availability}
    The datasets used to pretrain GALA-3D and MAE3D and to evaluate downstream segmentation are publicly available from their original sources \citep{merlin,kits23,kits19,Antonelli2022-qy,panorama,ichseg,li_state---art_2023,hematoma_jbhi,isles22}. Code and pretrained weights will be released upon publication at \url{https://github.com/aloyspx/GALA-3D}.

    \section*{Acknowledgments}
    This research was enabled in part by support provided by Calcul Qu\'ebec, Compute Ontario, SciNet, the BC DRI Group, and the Digital Research Alliance of Canada (alliancecan.ca).

    \bibliographystyle{cas-model2-names}
    \bibliography{preprint}

    \clearpage
    \appendix

    \section{Pretraining details}
    \label{app:pretraining}
    \setcounter{figure}{0}\setcounter{table}{0}

\begin{figure}[!bp]
        \centering
        \includegraphics[width=\columnwidth]{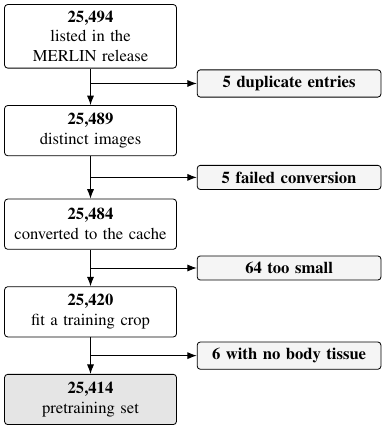}
        \caption{\textbf{MERLIN pretraining cohort.} Images remaining after each exclusion step.}
        \label{fig:cohort}
    \end{figure}

    \paragraph{Pretraining cohort} Figure~\ref{fig:cohort} shows the images removed at each step, and the remaining 25,414 images form the pretraining set.

    \paragraph{Hyperparameters} Table~\ref{tab:gala-hparams} gives the values of the symbols introduced in Section~\ref{sec:methodology}. Because pretraining volumes are stored on a $1\,\mathrm{mm}$ isotropic grid, source voxels correspond to millimetres. With $p=8$, a global view yields $N_g=12^3=1728$ patch tokens and a local view $N_l=8^3=512$, and the mask ratio leaves 86 visible and 1642 masked tokens per global view, of which 82 are queried. The SIGReg integral is evaluated with the trapezoidal rule, exploiting the even integrand. Operation counts (Section~\ref{sec:compute}) are measured with the PyTorch FLOP counter on one training step for one image, forward and backward, counting the backward pass of attention as twice its forward pass. Encoder sizes are listed in Table~\ref{tab:model-capacity}.

    \begin{table}[!tp]
        \caption{\textbf{GALA-3D pretraining hyperparameters.} FOV: field of view, in source voxels. The base learning rate is scaled inversely with encoder width.}
        \label{tab:gala-hparams}
        \vspace{5pt}
        \centering
        \footnotesize
        \begin{adjustbox}{max width=\columnwidth}
            % Hand-written table, moved out of main.tex on 2026-11-17. Edit here.
\begin{tabular}{llc}
    \toprule
    Symbol & Description & Value \\
    \midrule
    \multicolumn{3}{l}{\emph{View sampling}} \\
    --- & Coarse body mask stride & 8 \\
    --- & Parent edge & 128--144 \\
    --- & Parent candidates scored (at most) & 40 \\
    $\tau_P$ & Parent occupancy threshold & 0.5 \\
    $V_g$, $V_l$ & Global/local views per image & 2, 2 \\
    --- & Global FOV $\rightarrow$ input & $[96,128]\rightarrow96^3$ \\
    --- & Local FOV $\rightarrow$ input & $[64,85]\rightarrow64^3$ \\
    $\tau_F$ & View foreground occupancy & 0.3 \\
    $\tau_O$ & Local--global overlap & 0.3 \\
    \multicolumn{3}{l}{\emph{Architecture}} \\
    $p$ & Patch edge & 8 \\
    --- & Predictor blocks & 6 \\
    $g_{\psi}$, $h_{\xi}$ & Projection head layers & 3 \\
    $g_{\psi}$, $h_{\xi}$ & Projection head hidden dim & 2048 \\
    $k$ & Projection dimension & 128 \\
    \multicolumn{3}{l}{\emph{Objective}} \\
    $\lambda_g$, $\lambda_d$ & SIGReg weights & 0.05, 0.05 \\
    $J$ & SIGReg directions & 1024 \\
    $T$ & SIGReg integration bound & 5 \\
    --- & SIGReg trapezoidal knots on $[0,T]$ & 17 \\
    $\rho$ & Mask ratio & 0.95 \\
    $q$ & Query fraction & 0.05 \\
    $S$ & Dense SIGReg samples & 4 \\
    \multicolumn{3}{l}{\emph{Optimization}} \\
    $B$ & Distinct images per batch & 64 \\
    --- & Optimizer updates & 100,000 \\
    --- & Optimizer & AdamW \\
    --- & Weight decay & 0.05 \\
    $\beta$ & AdamW coefficients & (0.9, 0.95) \\
    --- & Base learning rate, ViT-S/8 & $5\times10^{-4}$ \\
    --- & Base learning rate, ViT-B/8 & $2.5\times10^{-4}$ \\
    --- & Linear warm-up & First 1\% of updates \\
    --- & Schedule after warm-up & Cosine decay to $10^{-3}$ of the base learning rate \\
    --- & Self-supervised gradient norm clip & 3 \\
    --- & Precision & Mixed bfloat16 \\
    \bottomrule
\end{tabular}

        \end{adjustbox}
    \end{table}

    \paragraph{Augmentation} Table~\ref{tab:augmentation} lists the transforms, which are sampled independently per view. Cropping, rotation, and resizing are fused into a single interpolation of the raw HU data. After clipping and scaling, each view is normalized using foreground moments computed from the full-image histogram under the selected window.

    \begin{table}[!tp]
        \caption{\textbf{Pretraining augmentations.} Noise and gamma are mutually exclusive: one of the two is applied with probability 0.5.}
        \label{tab:augmentation}
        \vspace{5pt}
        \centering
        \footnotesize
        \setlength{\tabcolsep}{3pt}
        \begin{adjustbox}{max width=\columnwidth}
            % Hand-written table, moved out of main.tex on 2026-11-17. Edit here.
\begin{tabular}{lc>{\raggedright\arraybackslash}p{4.0cm}}
    \toprule
    Transform & Probability & Setting \\
    \midrule
    Rotation & 1 & Continuous, $[-30^\circ,30^\circ]$ about the superior--inferior axis \\
    Flips & 0.3 each & Superior, anterior, and lateral, independently \\
    $90^\circ$ rotation & 0.3 & Axial plane \\
    Jittered clinical window & 0.8 & Soft-tissue, broad-lung, or dense-bone family (0.6, 0.2, 0.2), with level and width jittered subject to coverage, local-contrast, and saturation checks, otherwise the base window of the family or the soft-tissue window \\
    Fixed window & 0.2 & Soft tissue, $[-175,250]$\,HU \\
    Gaussian noise & $0.5\times0.4$ & Standard deviation 0.05 \\
    Gamma & $0.5\times0.6$ & Range-preserving, $\gamma\in[0.5,2]$ \\
    Blur or sharpen & 0.25 & Equal probability, independent of noise and gamma \\
    \bottomrule
\end{tabular}

        \end{adjustbox}
    \end{table}

    \paragraph{MAE3D control} MAE3D is pretrained on the same cohort with the same encoders and the optimization settings of Table~\ref{tab:gala-hparams}, for 100,000 optimizer updates. Following the MAE recipe \citep{he_masked_2021}, it uses a mask ratio of 0.75 and one random crop per image, without further augmentation.

    \section{Fine-tuning details}
    \label{app:finetuning}
    \setcounter{figure}{0}\setcounter{table}{0}

    Table~\ref{tab:finetuning} lists the fine-tuning settings. KiTS23, MSD Pancreas, and PANORAMA have nested targets (kidney $\supset$ masses $\supset$ tumour, and pancreas $\supset$ tumour), following the hierarchical evaluation classes of KiTS \citep{kits23}, and the remaining datasets have mutually exclusive classes. For PANORAMA, the pancreas comprises the parenchyma, the pancreatic duct, and the tumour, and the veins, arteries, and common bile duct, which the dataset also annotates, are treated as background.

    \begin{table}[!tp]
        \caption{\textbf{Fine-tuning settings.} Each combination of method, dataset, fold, and label budget is a single fine-tuning run with a fixed random seed.}
        \label{tab:finetuning}
        \vspace{5pt}
        \centering
        \footnotesize
        \setlength{\tabcolsep}{3pt}
        \begin{adjustbox}{max width=\columnwidth}
            \begin{tabular}{ll>{\raggedright\arraybackslash}p{3.7cm}}
    \toprule
    Group & Setting & Value \\
    \midrule
    Sampling & Crops per update & Two $96^3$ crops from one volume \\
     & Crop centres & Class-balanced (nested targets: foreground and background) \\
    \addlinespace
    Loss & Nested targets & Region-based multi-label, one sigmoid output per region, Dice + binary cross-entropy \\
     & Exclusive classes & Softmax output, Dice + cross-entropy \\
    \addlinespace
    Optimization & Optimizer & AdamW, weight decay $10^{-5}$ \\
     & Schedule & Cosine, linear warm-up over the first 1\% of updates \\
    \addlinespace
    Updates & One case and 1\% & 50,000, validation every 1,000 \\
     & 10\% & 100,000, validation every 1,000 \\
     & 100\% & 250,000, validation every 5,000 \\
     & Early stopping & No improvement in validation Dice for five consecutive validations \\
    \addlinespace
    Augmentation & Flips & Each spatial axis, probability 0.2 \\
     & $90^\circ$ rotation & Probability 0.2 \\
     & Intensity scaling & Factor within $\pm0.1$, probability 0.1 \\
     & Intensity shift & Offset within $\pm0.1$, probability 0.1 \\
    \addlinespace
    Testing & Checkpoint & Best validation Dice \\
     & Inference & Sliding window, 50\% overlap \\
    \bottomrule
\end{tabular}
        \end{adjustbox}
    \end{table}

    All recipe choices were made on validation data only. Fine-tuning learning rates were selected for each method by fine-tuning on KiTS23 at several label fractions and on a subset of folds and comparing validation Dice, without using the test folds, from candidates that generally included $1\times10^{-4}$, $3\times10^{-5}$, and $1\times10^{-5}$, as well as rates reported in their source publication. We selected $3\times10^{-5}$ for GALA-3D, Hi-End-MAE, VoCo, Curia-2, and UNETR-B, and $1\times10^{-5}$ for 3DINO, and MAE3D uses the same rate as GALA-3D. CT-FM, Swin-SSL, Models Genesis, UNETR-S, and Swin UNETR were kept at the shared $1\times10^{-4}$, as $3\times10^{-5}$ gave no consistent gain. The selected learning rate was then applied to all datasets and label budgets without further tuning.

    \section{Pretraining data of the compared models}
    \label{app:overlap}
    \setcounter{figure}{0}\setcounter{table}{0}

    \begin{table*}[!tp]
        \caption{\textbf{Pretraining data of the compared models, their documented overlap with the evaluation datasets, and their exposure to head CT.} Overlap was determined from each model's published list of pretraining datasets, including supplementary material and data-selection code. H\&N: head-and-neck cancer CT. ---: not reported as a number of volumes.}
        \label{tab:overlap}
        \vspace{5pt}
        \centering
        \footnotesize
        \setlength{\tabcolsep}{6pt}
        \begin{adjustbox}{max width=\textwidth}
            % Hand-written table, moved out of main.tex on 2026-11-17. Edit here.
\begin{tabular}{lrlll}
    \toprule
    Model & Volumes & Pretraining data & Overlap & Head CT in pretraining \\
    \midrule
    GALA-3D & 25,414 & MERLIN, abdominal CT & None & None \\
    Models Genesis & 623 & LUNA16, chest CT & None & None \\
    Swin-SSL & 5,050 & Public chest, head-and-neck, and colonography CT & None & H\&N (1.3k) \\
    VoCo-B/L & 160k & PreCT-160K, public CT & KiTS, MSD, PANORAMA & H\&N (4.3k) \\
    Hi-End-MAE & 10k & Public CT, incl.\ AbdomenAtlas~1.0 & KiTS, MSD, PANORAMA (MSD cases) & H\&N (42) \\
    3DINO & 100k & CT, MRI, and PET, 35 studies & KiTS (C4KC-KiTS), MSICH (via RSNA) & Yes (16k) \\
    CT-FM & 148k & CT, 69 IDC cohorts & KiTS (C4KC-KiTS) & None listed \\
    Curia-2 & --- & Private hospital CT and MRI & None & Yes, unquantified \\
    \bottomrule
\end{tabular}

        \end{adjustbox}
    \end{table*}

    The sources of Table~\ref{tab:overlap} are given below for each model. Overlap was determined from published dataset lists and was not audited at the case level. Prior exposure is self-supervised and does not involve our segmentation labels.

    \begin{itemize}
        \item \textbf{GALA-3D and MAE3D:} MERLIN \citep{merlin}, which contains no head images and is disjoint from every evaluation dataset.
        \item \textbf{Models Genesis:} LUNA16 chest CT, with no overlap or head CT \citep{modelgenesis}.
        \item \textbf{Swin-SSL:} chest, head-and-neck, and CT colonography images, including 1,287 head-and-neck volumes, with no overlap \citep{swinunetr}.
        \item \textbf{VoCo:} PreCT-160K lists 2,238 PANORAMA volumes and a pooled set of public tumour datasets that includes the MSD and KiTS. It also includes about 4,300 volumes from five head-and-neck cohorts \citep{voco_v2}.
        \item \textbf{Hi-End-MAE:} AbdomenAtlas~1.0 \citep{abdomenatlas,touchstone} and AbdomenCT-1K \citep{abdomenct1k} aggregate public abdominal CT benchmarks including KiTS and the MSD CT tasks. The corpus also includes 42 head-and-neck images from HaN-Seg \citep{hi_end_mae}.
        \item \textbf{3DINO:} includes C4KC-KiTS, whose cases form part of KiTS23, and brain CT from OASIS-3, an internal acute-stroke cohort, and a random half of the RSNA 2019 Intracranial Hemorrhage Detection dataset \citep{3dino,flanders_construction_2020}. MSICH was annotated from RSNA 2019 \citep{ichseg}, so 3DINO has likely seen part of the MSICH images.
        \item \textbf{CT-FM:} includes C4KC-KiTS, and its 69 cohorts include no head collection \citep{ctfm}.
        \item \textbf{Curia-2:} private clinical CT and MRI from one hospital, including examinations of the head, which were defaced before use \citep{curia2}.
    \end{itemize}

    Through the MSD pancreas task, VoCo and Hi-End-MAE have also seen the 97 PANORAMA cases in our set that originate from it (Section~\ref{sec:datasets}). The head-and-neck cohorts image the head and neck for cancer treatment rather than the brain. No corpus lists INSTANCE22 or ISLES22.

    \section{Domain-matched pretraining on head CT}
    \label{app:headct}
    \setcounter{figure}{0}\setcounter{table}{0}

    \begin{table*}[!b]
        \caption{\textbf{Domain-matched pretraining on head CT.} Held-out-test mean DSC (\%) over five folds on the two head-CT datasets. GALA-S, head CT is GALA-S pretrained on RSNA 2019 head CT with MSICH patients removed, with the objective and hyperparameters of GALA-S and head-CT intensity windows. Head CT: head CT in pretraining (Table~\ref{tab:overlap}). Avg: mean over the seven settings. Within each column, bold marks the best method and underline the second best.}
        \label{tab:headct}
        \vspace{5pt}
        \centering
        \footnotesize
        \setlength{\tabcolsep}{5pt}
        \begin{adjustbox}{max width=\textwidth}
            % Generated by downstream/NAB/plot/make_tex_tables.py; do not edit by hand.
\begin{tabular}{lc cccc ccc c}
    \toprule
    & & \multicolumn{4}{c}{MSICH} & \multicolumn{3}{c}{INSTANCE22} & \\
    \cmidrule(lr){3-6}\cmidrule(lr){7-9}
    Method & Head CT & 1 case & 1\% & 10\% & 100\% & 1 case & 10\% & 100\% & Avg \\
    \midrule
    UNETR-S (scratch) &  & 30.2 & 41.1 & 58.0 & 71.8 & 44.2 & 61.9 & 70.6 & 54.0 \\
    Models Genesis &  & 36.5 & 44.6 & 59.9 & 70.7 & 50.6 & 67.0 & 73.5 & 57.5 \\
    Swin-SSL &  & 34.7 & 43.5 & 60.8 & 72.7 & 49.0 & 65.5 & 73.4 & 57.1 \\
    VoCo-B &  & 33.2 & 44.2 & 61.2 & 73.0 & 47.5 & 66.0 & 72.9 & 56.9 \\
    VoCo-L &  & 34.3 & 43.5 & 62.3 & \underline{73.3} & 48.1 & 66.9 & 72.4 & 57.3 \\
    Hi-End-MAE &  & 37.7 & 47.9 & 63.5 & 72.9 & 47.1 & 64.7 & 71.5 & 57.9 \\
    3DINO & $\checkmark$ & 34.8 & 45.0 & 58.7 & 70.9 & 47.5 & 62.7 & 70.0 & 55.7 \\
    CT-FM &  & 34.6 & 46.5 & 62.1 & 72.1 & 48.1 & \textbf{68.7} & 72.6 & 57.8 \\
    Curia-2 & $\checkmark$ & 35.3 & \underline{48.2} & \underline{64.8} & \textbf{73.8} & 46.2 & 65.9 & \underline{74.1} & 58.3 \\
    \midrule
    GALA-S &  & \underline{37.8} & 47.9 & 62.4 & 73.2 & \underline{51.8} & 67.3 & \textbf{74.4} & \underline{59.3} \\
    \midrule
    GALA-S, head CT & $\checkmark$ & \textbf{40.1} & \textbf{52.9} & \textbf{66.3} & 72.9 & \textbf{52.5} & \underline{67.8} & 73.3 & \textbf{60.8} \\
    \bottomrule
\end{tabular}

        \end{adjustbox}
    \end{table*}

    \paragraph{Setup} To test whether the head-CT results of GALA-3D are limited by its abdominal pretraining data rather than by the method, GALA-S was pretrained a second time on head CT. The cohort consists of the 24,458 images of the RSNA 2019 Intracranial Hemorrhage Detection dataset \citep{flanders_construction_2020} that remain after removing every image of the MSICH patients, since MSICH was annotated from RSNA 2019 \citep{ichseg}. It is therefore of similar size to MERLIN (25,414 images), disjoint from both evaluation datasets, and close to the domain of MSICH. Only the data and the clinical windows used to generate the views differ from GALA-S (Section~\ref{sec:implementation-details}), with three head-CT windows (brain L40/W80, subdural L80/W200, and soft tissue L40/W380\,HU) in place of the abdominal ones, and fine-tuning follows the main protocol.

    \paragraph{Results} On MSICH, where GALA-S trailed the strongest external models with 1\% and 10\% of the labels, head-CT pretraining makes GALA-S the best method at every limited budget (Table~\ref{tab:headct}). It reaches 40.1, 52.9, and 66.3 DSC with one labelled case, 1\%, and 10\% of the labels, an improvement of 2.3, 5.0, and 3.9 points over GALA-S pretrained on MERLIN, and it exceeds that model on all 15 folds. It also exceeds the strongest external model of each budget on 13 of these 15 folds. Two external models were pretrained on intracranial head CT (Table~\ref{tab:overlap}). 3DINO, whose 16k head-CT volumes include part of the RSNA 2019 images from which MSICH was drawn, trails head-CT GALA-S by 5.3 to 7.9 points at these budgets. Curia-2, pretrained on clinical CT and MRI that include the head, trails it on 14 of the 15 folds. The gain decreases as labels increase, and with the full training pool the two GALA-S models are on par (72.9 and 73.2 DSC). On INSTANCE22, where GALA-S was already competitive, head-CT pretraining changes its DSC by $+0.7$, $+0.5$, and $-1.1$ points, and it remains the best method with one labelled case.

    \paragraph{Interpretation} These results indicate that the remaining gap of GALA-3D on MSICH mainly reflects its pretraining data rather than the objective. With a cohort of similar size from the target anatomy and only the intensity windows adapted, GALA-S becomes the strongest initialization on MSICH whenever labels are limited.

    \section{Additional downstream results}
    \label{app:results}
    \setcounter{figure}{0}\setcounter{table}{0}

\begin{figure*}[!tp]
        \centering
        \includegraphics[width=\textwidth]{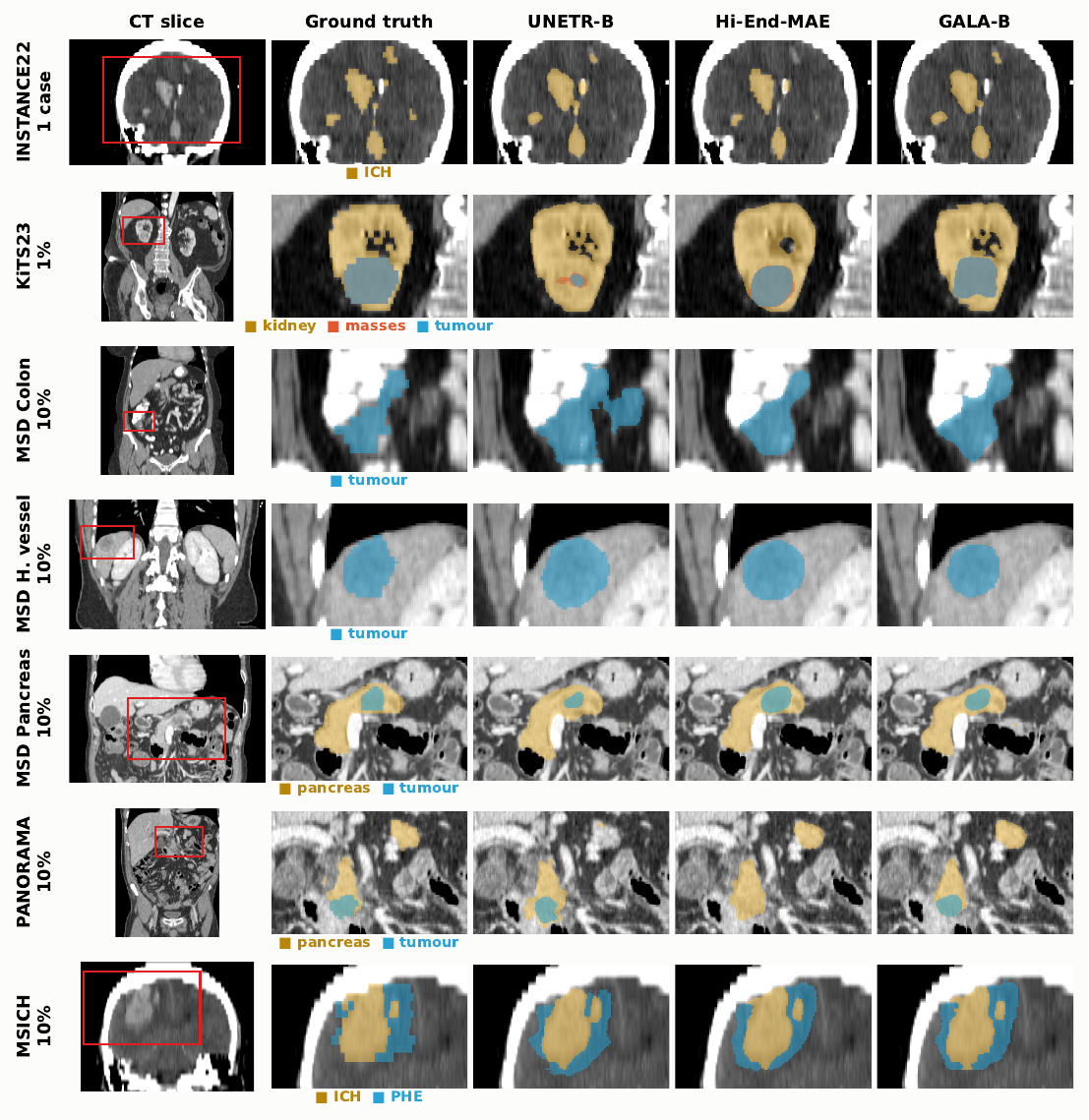}
        \caption{\textbf{Coronal view of the test cases of Figure~\ref{fig:qualitative}.} Same cases, models, and label budgets as Figure~\ref{fig:qualitative}, shown in a coronal plane through the same lesion, selected as for the axial figure. Left: full slice, with the red box marking the region shown in the other columns.}
        \label{fig:qualitative-coronal}
    \end{figure*}

    Tables~\ref{tab:stats-vitb8} and~\ref{tab:stats-vits8} give the paired comparisons of GALA-B and GALA-S with every other method in DSC. Table~\ref{tab:class-margins} gives the margin of GALA-B over the strongest external model for each target class of the abdominal datasets. Tables~\ref{tab:nsd-vitb8} and~\ref{tab:nsd-tolerance} report boundary agreement in NSD. Figure~\ref{fig:qualitative-coronal} shows the cases of Figure~\ref{fig:qualitative} in the coronal plane.

    \begin{table*}[!tp]
        \centering
        \footnotesize
        \begin{minipage}[t]{0.485\textwidth}
            \caption{\textbf{Paired comparison of GALA-B with training from scratch, every external model, and GALA-S on the seven CT datasets (five at 1\%).} Mean difference in held-out-test DSC (points) over (dataset, fold) pairs. Two-sided Wilcoxon signed-rank test, Holm-corrected over all 48 comparisons: $^{*}p<0.05$, $^{**}p<0.01$. Params: parameters of each comparator's fine-tuned network, encoder and decoder (GALA-B: 99M).}
            \label{tab:stats-vitb8}
            \vspace{5pt}
            \centering
            \setlength{\tabcolsep}{2pt}
            \begin{adjustbox}{max width=\linewidth}
                % Generated by downstream/NAB/plot/make_paper_stats.py; do not edit by hand.
\begin{tabular}{lrrrrr}
  \toprule
  Comparator & Params & 1 case & 1\% & 10\% & 100\% \\
  \midrule
  UNETR-S & 33M & +15.8$^{**}$ & +23.9$^{**}$ & +17.2$^{**}$ & +7.5$^{**}$ \\
  UNETR-B & 99M & +15.1$^{**}$ & +23.0$^{**}$ & +16.8$^{**}$ & +7.5$^{**}$ \\
  Swin UNETR & 62M & +15.4$^{**}$ & +22.8$^{**}$ & +15.1$^{**}$ & +6.2$^{**}$ \\
  Models Genesis & 19M & +13.7$^{**}$ & +21.3$^{**}$ & +11.9$^{**}$ & +7.8$^{**}$ \\
  Swin-SSL & 62M & +12.5$^{**}$ & +20.0$^{**}$ & +13.8$^{**}$ & +6.6$^{**}$ \\
  VoCo-B & 73M & +4.2$^{**}$ & +7.1$^{**}$ & +3.2$^{**}$ & +1.3$^{**}$ \\
  VoCo-L & 290M & +4.7$^{**}$ & +6.9$^{**}$ & +2.9$^{**}$ & +0.9 \\
  Hi-End-MAE & 361M & +2.3$^{*}$ & +3.1$^{**}$ & +2.9$^{**}$ & +2.6$^{**}$ \\
  3DINO & 322M & +9.1$^{**}$ & +12.8$^{**}$ & +8.7$^{**}$ & +5.9$^{**}$ \\
  CT-FM & 87M & +12.0$^{**}$ & +15.9$^{**}$ & +7.8$^{**}$ & +4.2$^{**}$ \\
  Curia-2 & 100M & +4.1$^{**}$ & +5.4$^{**}$ & +3.5$^{**}$ & +1.4 \\
  GALA-S & 33M & -0.0 & +1.9$^{*}$ & +1.2$^{**}$ & +0.1 \\
  \bottomrule
\end{tabular}

            \end{adjustbox}
        \end{minipage}\hfill
        \begin{minipage}[t]{0.485\textwidth}
            \caption{\textbf{Paired comparison of GALA-S with every other method in DSC.} Mean difference in held-out-test DSC (points) over (dataset, fold) pairs on the seven CT datasets (five at 1\%), with tests as in Table~\ref{tab:stats-vitb8}. Holm correction is applied within this table, so the GALA-B row can differ in significance from the corresponding row of Table~\ref{tab:stats-vitb8}. Params as in Table~\ref{tab:stats-vitb8} (GALA-S: 33M).}
            \label{tab:stats-vits8}
            \vspace{5pt}
            \centering
            \setlength{\tabcolsep}{2pt}
            \begin{adjustbox}{max width=\linewidth}
                % Generated by downstream/NAB/plot/make_paper_stats.py; do not edit by hand.
\begin{tabular}{lrrrrr}
  \toprule
  Comparator & Params & 1 case & 1\% & 10\% & 100\% \\
  \midrule
  UNETR-S & 33M & +15.8$^{**}$ & +22.0$^{**}$ & +16.0$^{**}$ & +7.4$^{**}$ \\
  UNETR-B & 99M & +15.1$^{**}$ & +21.1$^{**}$ & +15.6$^{**}$ & +7.5$^{**}$ \\
  Swin UNETR & 62M & +15.4$^{**}$ & +20.9$^{**}$ & +13.8$^{**}$ & +6.1$^{**}$ \\
  Models Genesis & 19M & +13.7$^{**}$ & +19.4$^{**}$ & +10.7$^{**}$ & +7.7$^{**}$ \\
  Swin-SSL & 62M & +12.5$^{**}$ & +18.1$^{**}$ & +12.6$^{**}$ & +6.6$^{**}$ \\
  VoCo-B & 73M & +4.3$^{**}$ & +5.2$^{**}$ & +1.9$^{**}$ & +1.3$^{*}$ \\
  VoCo-L & 290M & +4.7$^{**}$ & +5.0$^{**}$ & +1.7$^{**}$ & +0.8 \\
  Hi-End-MAE & 361M & +2.3$^{*}$ & +1.2 & +1.7$^{*}$ & +2.5$^{**}$ \\
  3DINO & 322M & +9.1$^{**}$ & +10.9$^{**}$ & +7.5$^{**}$ & +5.8$^{**}$ \\
  CT-FM & 87M & +12.0$^{**}$ & +14.0$^{**}$ & +6.5$^{**}$ & +4.1$^{**}$ \\
  Curia-2 & 100M & +4.1$^{**}$ & +3.5$^{*}$ & +2.3$^{*}$ & +1.3 \\
  GALA-B & 99M & +0.0 & -1.9$^{*}$ & -1.2$^{*}$ & -0.1 \\
  \bottomrule
\end{tabular}

            \end{adjustbox}
        \end{minipage}
    \end{table*}

    \begin{table*}[!tp]
        \centering
        \footnotesize
        \begin{minipage}[t]{0.485\textwidth}
            \caption{\textbf{Margin of GALA-B over the strongest external model for each target class of the abdominal CT datasets.} Difference in held-out-test DSC (points, mean over five folds) between GALA-B and the best of the eight external models for that class and label budget. --: 1\% equals one labelled case on MSD Colon.}
            \label{tab:class-margins}
            \vspace{5pt}
            \centering
            % Generated by downstream/NAB/plot/make_appendix_tables.py; do not edit by hand.
\begin{tabular}{lcccc}
  \toprule
  Class & 1 case & 1\% & 10\% & 100\% \\
  \midrule
  \multicolumn{5}{l}{\emph{Lesions}} \\
  KiTS23 masses & $+2.7$ & $+6.5$ & $+1.4$ & $+0.7$ \\
  KiTS23 tumour & $+4.0$ & $+7.4$ & $+3.3$ & $+1.8$ \\
  MSD H.\,vessel tumour & $+3.4$ & $+4.6$ & $-0.4$ & $+0.2$ \\
  MSD Pancreas tumour & $+3.9$ & $+3.2$ & $+2.7$ & 0.0 \\
  PANORAMA PDAC & $+4.6$ & $+6.1$ & $+2.3$ & $+0.2$ \\
  MSD Colon tumour & $+2.4$ & -- & $+7.4$ & $+1.9$ \\
  \addlinespace
  \multicolumn{5}{l}{\emph{Organs and vessels}} \\
  KiTS23 kidney & $+1.7$ & $+3.1$ & 0.0 & $+0.3$ \\
  MSD H.\,vessel vessels & $-0.1$ & 0.0 & $-0.1$ & $+0.1$ \\
  MSD Pancreas pancreas & $+0.4$ & $-1.5$ & 0.0 & $+0.2$ \\
  PANORAMA pancreas & $-2.0$ & $-1.8$ & $+0.6$ & $-0.3$ \\
  \bottomrule
\end{tabular}

        \end{minipage}\hfill
        \begin{minipage}[t]{0.485\textwidth}
            \caption{\textbf{Paired comparison of GALA-B with every other method in NSD.} Mean difference in held-out-test NSD at a tolerance of $2\,\mathrm{mm}$ (points) over (dataset, fold) pairs on the seven CT datasets (five at 1\%). Two-sided Wilcoxon signed-rank test, Holm-corrected over all 48 comparisons: $^{*}p<0.05$, $^{**}p<0.01$.}
            \label{tab:nsd-vitb8}
            \vspace{5pt}
            \centering
            \setlength{\tabcolsep}{2pt}
            \begin{adjustbox}{max width=\linewidth}
                % Generated by downstream/NAB/plot/make_nsd_tables.py; do not edit by hand.
\begin{tabular}{lrrrr}
  \toprule
  Comparator & 1 case & 1\% & 10\% & 100\% \\
  \midrule
  UNETR-S & +13.5$^{**}$ & +20.7$^{**}$ & +17.9$^{**}$ & +8.7$^{**}$ \\
  UNETR-B & +12.7$^{**}$ & +20.3$^{**}$ & +17.8$^{**}$ & +8.6$^{**}$ \\
  Swin UNETR & +13.0$^{**}$ & +19.7$^{**}$ & +15.7$^{**}$ & +7.4$^{**}$ \\
  Models Genesis & +10.7$^{**}$ & +18.1$^{**}$ & +11.4$^{**}$ & +8.5$^{**}$ \\
  Swin-SSL & +10.6$^{**}$ & +17.4$^{**}$ & +14.4$^{**}$ & +7.8$^{**}$ \\
  VoCo-B & +3.7$^{**}$ & +5.8$^{**}$ & +2.9$^{**}$ & +1.5$^{*}$ \\
  VoCo-L & +3.9$^{**}$ & +5.2$^{**}$ & +2.0$^{**}$ & +0.4 \\
  Hi-End-MAE & +2.1$^{*}$ & +2.8$^{**}$ & +3.5$^{**}$ & +3.0$^{**}$ \\
  3DINO & +8.6$^{**}$ & +12.7$^{**}$ & +11.1$^{**}$ & +7.2$^{**}$ \\
  CT-FM & +10.1$^{**}$ & +13.8$^{**}$ & +8.4$^{**}$ & +4.9$^{**}$ \\
  Curia-2 & +3.6$^{**}$ & +4.3$^{**}$ & +3.3$^{**}$ & +1.3 \\
  GALA-S & -0.2 & +2.1$^{*}$ & +1.2$^{*}$ & -0.2 \\
  \bottomrule
\end{tabular}

            \end{adjustbox}
        \end{minipage}
    \end{table*}

    \begin{table*}[!tp]
        \centering
        \footnotesize
        \begin{minipage}[t]{0.7\textwidth}
            \caption{\textbf{Sensitivity of the NSD comparison to the tolerance.} Mean NSD margin (points) of GALA-B over the eight external models at each tolerance and label budget, followed by the number of external models GALA-B exceeds and the number of these differences that are significant (Holm-corrected Wilcoxon signed-rank tests over the 32 external-model comparisons of each tolerance, $p<0.05$). At 2\,mm, these counts match those of Table~\ref{tab:nsd-vitb8}, where the same comparisons are Holm-corrected over all 48 comparisons.}
            \label{tab:nsd-tolerance}
            \vspace{5pt}
            \centering
            \setlength{\tabcolsep}{3pt}
            \begin{adjustbox}{max width=\linewidth}
                % Generated by downstream/NAB/plot/make_nsd_tables.py; do not edit by hand.
\begin{tabular}{lcccc}
  \toprule
  Tolerance & 1 case & 1\% & 10\% & 100\% \\
  \midrule
  1\,mm & +4.6 (8/8, 8/8) & +7.4 (8/8, 8/8) & +5.4 (8/8, 6/8) & +3.6 (7/8, 5/8) \\
  2\,mm & +6.6 (8/8, 8/8) & +10.0 (8/8, 8/8) & +7.1 (8/8, 8/8) & +4.3 (8/8, 6/8) \\
  5\,mm & +10.0 (8/8, 8/8) & +13.2 (8/8, 8/8) & +8.6 (8/8, 8/8) & +4.9 (8/8, 8/8) \\
  \bottomrule
\end{tabular}

            \end{adjustbox}
        \end{minipage}
    \end{table*}

\end{document}